\documentclass[lettersize,journal]{IEEEtran}
\usepackage{amsmath,amsfonts}

\usepackage{algorithm}
\usepackage{array}
\usepackage{subcaption}
\usepackage{textcomp}
\usepackage{stfloats}
\usepackage{verbatim}
\usepackage{graphicx}
\usepackage{cite}

\usepackage{hyperref}
\usepackage{moreverb,url}

\usepackage{subcaption}
\usepackage{multirow}
\usepackage{multicol}
\usepackage{enumitem}

\usepackage{blindtext}
\usepackage{amssymb}
\usepackage{tabularx}
\usepackage{algpseudocode}
\usepackage[table]{xcolor}
\usepackage{color}
\usepackage{siunitx}
\DeclareSIUnit\pixel{px}
\usepackage{tikz}

\usepackage{xcolor}
\usepackage{orcidlink}

\definecolor{mygreen}{RGB}{150,0,0}
\definecolor{myred}{RGB}{0,0,100}
\definecolor{myblue}{RGB}{0,60,0}
\definecolor{myyellow}{RGB}{150, 100, 0}

\usepackage[acronym]{glossaries}
\glsdisablehyper
\newacronym{mse}{MSE}{Mean Squared Error}
\newacronym{bce}{BCE}{Binary Cross Entropy}
\newacronym{svd}{SVD}{Singular Value Decomposition}
\newacronym{icp}{ICP}{Iterative Closest Point}
\newacronym{slam}{SLAM}{Simultaneous Localisation and Mapping}
\newacronym{dl}{DL}{Deep Learning}
\newacronym{gat}{GAT}{Graph Attention Network}
\newacronym{gnn}{GNN}{Graph Neural Network}
\newacronym{rte}{RTE}{Relative Translation Error}
\newacronym{rre}{RRE}{Relative Rotation Error}
\newacronym{nar}{NAR}{Neural Algorithmic Reasoning}
\newacronym{cce}{CCE}{Categorical Cross Entropy}
\newacronym{mpnn}{MPNN}{Message-Passing Neural Network}
\newacronym{mlp}{MLP}{Multilayer Perceptron}

\newif\ifanonymous
\anonymousfalse  

\usepackage{cleveref}

\crefname{table}{Table}{Tables}
\crefname{figure}{Figure}{Figures}
\crefname{section}{Section}{Sections}

\crefname{equation}{Equation}{Equations}

\begin{document}

\title{NAR-*ICP: Neural Execution of Classical ICP-based Point Cloud Registration Algorithms}

\ifanonymous
\author{\textbf{Anonymous Authors}, Paper under double-blind review
\thanks{
\newline
The code for NAR-*ICP and the dataset generation will be made available upon acceptance.}
}
\else

\author{Efimia Panagiotaki~\IEEEmembership{Member,~IEEE,}
Daniele De Martini~\IEEEmembership{Member,~IEEE,}
Lars Kunze~\IEEEmembership{Member,~IEEE,} \\
Paul Newman~\IEEEmembership{Fellow,~IEEE,} and 
Petar Veli\v{c}kovi\'c~\IEEEmembership{Member,~IEEE}
\thanks{ The authors Efimia Panagiotaki, Daniele De Martini, Lars Kunze, and Paul Newman are with the Oxford Robotics Institute (ORI), University of Oxford, UK. Lars Kunze is additionally with the Bristol Robotics Laboratory (BRL), University of the West of England, UK. Petar Veli\v{c}kovi\'c is with Google DeepMind and the University of Cambridge, UK.}
\thanks{Corresponding author: Efimia Panagiotaki, efimia@robots.ox.ac.uk}
}
\fi



\maketitle

\begin{abstract}
This study explores the intersection of neural networks and classical robotics algorithms through the \gls{nar} blueprint, enabling the training of neural networks to reason like classical robotics algorithms by learning to execute them. Algorithms are integral to robotics and safety-critical applications due to their predictable and consistent performance through logical and mathematical principles. In contrast, while neural networks are highly adaptable, handling complex, high-dimensional data and generalising across tasks, they often lack interpretability and transparency in their internal computations. To bridge the two, we propose a novel \gls{gnn}-based framework, NAR-*ICP, that learns the intermediate computations of classical ICP-based registration algorithms, extending the CLRS Benchmark. We evaluate our approach across real-world and synthetic datasets, demonstrating its flexibility in handling complex inputs, and its potential to be used within larger learning pipelines. Our method achieves superior performance compared to the baselines, even surpassing the algorithms it was trained on, further demonstrating its ability to generalise beyond the capabilities of traditional algorithms.




\end{abstract}
\begin{IEEEkeywords}
Robot learning, graph neural networks, robot localisation, registration, neural algorithmic reasoning
\end{IEEEkeywords}


\glsresetall


\section{Introduction \label{sec:intro}}
\IEEEPARstart{A}{lgorithms} serve as the foundation for tackling a wide range of robotics tasks, from path planning and optimisation to perception and control.
Classical robotics algorithms are valued for their inherent modularity, ensuring reliable and consistent task execution.
Their interpretable nature enables transparency and accountability in robotic operations, which are essential for understanding and validating the behaviour of robotic systems.
However, traditional algorithms require strictly pre-defined data specifications, limiting their ability to process raw inputs and handle complex datasets.

Neural networks, in contrast, excel in efficiently handling raw sensor inputs and demonstrate robustness to noisy and complex data.
By learning intricate patterns and dependencies within the input datasets, neural networks effectively generalise, performing reliably even in diverse and previously unseen scenarios.
However, they lack interpretability, which poses challenges for transparency and accountability in robotic operations.

Traditionally, neural networks have been trained using the ground truth directly as the supervisory signal, establishing direct mappings from raw inputs to the respective outputs without providing insights into their internal computational steps.
Despite their strong performance, these networks lack transparency into the models' decision-making processes, creating a significant gap in interpretability and reasoning. Aiming to combine the strengths of classical algorithms and neural networks, the \gls{nar} \cite{nar} blueprint proposes the integration of algorithmic computations within neural networks. Recent \gls{nar} works achieve this by training a network to robustly approximate an algorithm by teaching it to mimic its reasoning through supervision on the intermediate algorithmic steps and outputs \cite{dar, neuralexecution}.
This hybrid approach offers flexible, scalable, and robust performance in complex and diverse inputs, robustness to noise and variability, and interpretable behaviour for prior black-box models.
These features make this method well-suited for a wide range of tasks while bridging the gap between traditional algorithms and neural networks in robotics.

This work introduces a novel robot learning methodology, built on the \gls{nar} blueprint and adapted to classical robotics perception algorithms, with point cloud registration as the primary task.
We focus on \gls{icp}-based algorithms as they are a fundamental registration methodology, widely accepted and used in robotics \cite{yang20243dregistration30years}.
Their compositional structure and iterative nature make them a strong candidate for extending \gls{nar} in the robotics domain.
Here, we train \glspl{gnn} to emulate these traditional algorithms on various complex and real-world datasets, inheriting their interpretability within the networks, surpassing their performance, and overcoming key algorithmic limitations, such as the critical need for a strong initial guess.

Our key contributions are as follows:
\begin{enumerate}
\item A novel graph learning methodology, grounded in the \gls{nar} blueprint, designed to approximate the reasoning processes of point cloud registration algorithms and generalise robustly on real-world datasets.
\item An extension of \gls{nar}, introducing real-world data integration, 3D geometric reasoning through point cloud representations, learned termination and algorithmic convergence, flexible hint configurations, scalable training, and ground-truth optimisation.
\item A comprehensive evaluation of the neural execution of fundamental \gls{icp}-based algorithms on complex and noisy data, including point clouds with low overlap, demonstrating superior performance compared to the algorithms while overcoming their key limitations and enabling seamless integration into larger learning systems.
\item A thorough comparison with state-of-the-art learned registration benchmarks, which we adapt and optimise for object-based registration, achieving superior performance across both accuracy and runtime.
\end{enumerate}


To the best of our knowledge, this is the first work applying \gls{nar} to a fundamental robotics perception task and the first to apply \gls{nar} to a task of this complexity. 

The remainder of the paper is organised as follows: \cref{sec:litreview} frames this work in the literature, while \cref{sec:overview} gives an overview of the foundational frameworks and outlines our method. \Cref{sec:methodology} details the methodology and key contributions. \Cref{sec:experiments} describes the datasets, algorithms, benchmarks, and experimental setup, and \cref{sec:results} presents the experiments and results, showcasing the findings and discussing their significance. \Cref{sec:ablation} proposes ablations to examine the contributions of different components to our model, and \Cref{sec:learning_pipeline} discusses the integration of NAR-*ICP into learning pipelines. Finally, \cref{sec:conclusion} concludes the paper with a summary of our contributions, outlining potential directions for future research.


\section{Related Work \label{sec:litreview}}
This work intersects three key domains: robot learning, neural algorithmic reasoning, and point cloud registration. In this section, we provide a brief overview of relevant works within each field, highlighting those that have particularly influenced our approach. We also discuss how our proposed method differs from and builds upon existing research.

\subsection{Robot Learning} Recent advances in machine learning have significantly enhanced the robots' ability to perceive, navigate, and adapt to complex and dynamic environments. Reinforcement learning, for instance, enables robots to learn optimal policies through interaction with their environment and feedback-driven exploration \cite{rl1, rl2}. In contrast, imitation learning reduces the need for extensive trial-and-error training, relying mainly on learning directly from expert demonstrations \cite{imitation, interimitationlearning}. To improve efficiency, transfer learning allows knowledge gained in one task to be applied to others, minimising the need for retraining in each new application \cite{transferlearning}. Multi-task learning takes a complementary approach by enabling robots to learn several tasks simultaneously, sharing knowledge across to enhance overall performance \cite{multitasklearning}. More recently, foundation models have shown impressive generality across domains, leveraging extensive large-scale pre-training and model fine-tuning for a wide range of downstream tasks \cite{foundationmodels}. Additionally, \glspl{gnn} have introduced powerful inductive biases,  incorporating relational and structural priors, effectively enabling models to reason over spatial, temporal, and graph-structured data in complex robotic scenarios \cite{graphlearning}. 

Despite these advances, the foundations of robotics are deeply algorithmic. From perception modules, such as feature matching and registration, to planning and control, classical algorithms remain at the core of nearly every robotics pipeline. Yet, conventional learning approaches struggle to capture precise algorithmic computations within the networks. Neural Algorithmic Reasoning (NAR) \cite{nar} has emerged as a strong framework that trains neural networks to mimic the reasoning processes of classical algorithms, primarily those in the CLRS-30 Benchmark \cite{clrs, clrs_book}. The NAR blueprint presents a way to combine the flexibility of learning-based approaches with the reliability of algorithmic reasoning \cite{neuralexecution}. While NAR has primarily been studied in synthetic domains, its potential impact in robotics is evident. In this work, we present the first adaptation of NAR to a fundamental robotics perception task: point cloud registration. 

\subsection{Neural Algorithmic Reasoning} Recent studies propose the concept of algorithmic alignment, suggesting that aligning the learning process with the steps of a target algorithm facilitates optimisation \cite{Xu2020What}. \Gls{nar} \cite{nar} integrates neural networks, most prominently \glspl{gnn} \cite{neuralexecution, clrs}, with classical algorithms, without relying on large datasets. 
Unlike traditional learning frameworks, \gls{nar} models learn to execute intermediate algorithmic steps rather than directly mapping inputs to outputs. 
They leverage the flexibility of neural networks to capture intricate patterns and relationships in the input, while being trained to emulate algorithmic computations. 
A key advantage of \gls{nar} models is their potential for out-of-distribution (OOD) generalisation, as they rely on algorithmic priors to guide their predictions instead of relying on arbitrary data correlations, enabling generalisation beyond the training distributions \cite{causal, mahdavi2023towards, NARSupervision, minder2023salsaclrs}. 

This framework has been applied in various domains, including large language models \cite{transnar}, reinforcement learning \cite{XLVIN}, planning and decision-making \cite{planners, ContinuousPlanners}, and explainability \cite{explainability}. 
Beyond applications, recent works extend the \gls{nar} paradigm itself by enforcing discrete state transitions for improved neural algorithmic execution \cite{rodionov2025discrete}, exploring compositional dual reasoning \cite{dar}, investigating transfer learning \cite{transferNAR}, ensuring consistency with the Markov property during execution \cite{markovproperty}, and exploring learning approaches for problems with multiple valid solutions \cite{multiplecorrect}. Moreover, \gls{nar} has been evaluated and extended not only for polynomial-time-solvable tasks but also NP-hard/complete problems through the primal-dual paradigm \cite{he2025primaldual} and combinatorial optimisation \cite{combinatorialoptimisation}.

In this work, we extend the \gls{nar} framework and the CLRS-30 Benchmark \cite{clrs}, integrating complex multi-step point cloud registration algorithms, real-world datasets, and optimising the output using ground truth supervisory signals and a termination network. We evaluate our approach on various challenging synthetic and real-world datasets, and test our method as part of a larger learning system. 

\subsection{Point Cloud Registration} The \gls{icp} algorithm \cite{icp, icp2} is considered one of the fundamental robotics algorithms, widely used to solve rigid registration tasks \cite{icp_variants, icp_datasets}. \gls{icp} iteratively finds correspondences between two sets of points by matching closest pairs and estimating the relative transformation between them to eventually align them. Multiple methods have extended \gls{icp} to address different robotics challenges, such as handling noise and outliers in input datasets \cite{newicp, gicp}, dealing with non-rigid deformations \cite{nonrigid1, nonrigid2}, ensuring robustness through adaptive thresholding and frame-to-map registration \cite{Vizzo2023}, accelerating alignment \cite{fasticp, point-plane}, and its integration with other perception algorithms for autonomous navigation \cite{guadagnino2025kissslamsimplerobustaccurate, icpslam}. 

More recently, learned registration methods have extended the closest-correspondence and relative transformation principles of \gls{icp}. Approaches such as Deep Global Registration (DGR) \cite{dgr} built on FCGF \cite{fcgf}, Deep Closest Point (DCP) \cite{dcp}, and DeepICP \cite{deepicp} have shown promising results and are the most relevant to our approach. However, they primarily map inputs directly to outputs rather than learning the reasoning steps or following the algorithmic computations of \gls{icp} itself. Beyond \gls{icp}-based pipelines, recent state-of-the-art methods, such as Predator \cite{predator} and its successor GeoTransformer \cite{geotransformer}, have advanced large-scale, dense point cloud registration for point clouds with low overlap. Despite their strong performance, their reliance on dense point clouds leads to significant computational and memory costs. Aiming for lightweight registration, several methods, such as SEM-GAT \cite{semgat}, TMP \cite{tmp}, PADLoC \cite{padloc}, InstaLoc \cite{instaloc}, and SGPR \cite{Kong2020sgpr}, integrate semantics into the registration process and, in some cases, the loop closure pipeline, leveraging semantic labels to improve correspondence estimation. However, they still rely on dense point clouds to capture object-level geometry and, similar to all other learned registration pipelines, they are trained directly on ground-truth data, lacking transparency and interpretability in their internal reasoning.

In contrast, our method relies on smaller point clouds, optimised for algorithmic processing, reducing the overhead associated with the dense inputs required by most learning approaches. We use \gls{icp}-based algorithms as supervisory signals for training networks to emulate intermediate algorithmic computations. This explicitly grounds them in the underlying algorithmic reasoning, addressing the transparency and interpretability limitations in conventional end-to-end learning. In our evaluation, we compare our models against the algorithms used for supervision, as well as learned \gls{icp}-based methods and state-of-the-art approaches.

\section{Foundations and Method Overview \label{sec:overview}}
\renewcommand{\thefootnote}{\arabic{footnote}}

This section gives an overview of NAR-*ICP\footnote{The asterisk (*) in NAR-*ICP refers to the different ICP-based algorithms that are approximated using \gls{nar} in this paper.}, positioning it within the context of \gls{nar} and point cloud registration.

\subsection{Point Cloud Registration}
Point cloud registration algorithms take as input two scans captured at different timesteps $m$ and $n$, represented as $P_m= \{\mathbf{p}_i \mid \mathbf{p}_i \in \mathbb{R}^3\}$ and $P_n= \{\mathbf{p}_j \mid \mathbf{p}_j \in \mathbb{R}^3\}$ where each $\mathbf{p}_i$, $\mathbf{p}_j$ represents a point in three-dimensional space. We assume that $P_n$ is transformed from $P_m$ by a rigid transformation denoted by $[\mathbf{R}_{m,n} \mid \mathbf{t}_{m,n}]$, where $\mathbf{R}_{m,n} \in SO(3)$ is an orthogonal matrix representing the rotation of a point and $\mathbf{t}_{m,n} \in \mathbb{R}^3$ is a translation vector representing the displacement of the points in 3D space. 
\gls{icp}-based algorithms, here $\mathcal{A}$, aim to minimise the difference between $P_n$ and the transformed $P_m$ by iteratively minimising an error function $\mathrm{\mathbf{e}}^{(t)}$, following a two-phase process. First, they find correspondences $C^{(t)} = \{ (\mathbf{p}_i, \mathbf{p}_j) \mid \mathbf{p}_i \in P_m, \mathbf{p}_j \in P_n \}$ by matching each point in $P_n$ to a point in $P_m$. Then, they leverage these correspondences to estimate the relative transformation that aligns the point clouds. This iterative process continues until the error criterion $\mathrm{\mathbf{e}}$ is minimised or until the algorithm reaches a predefined maximum number of iterations. The general structure of ICP-based algorithms is detailed in the pseudocode in \Cref{algo:icp}.

\begin{algorithm}[]
\caption{ICP-based Algorithms \label{algo:icp}} 
\begin{tikzpicture}[baseline]
    \draw[line width=2pt, orange!80!black] (0,0) -- (0.2,0);
\end{tikzpicture}
\small \textit{Phase 1}
\begin{tikzpicture}[baseline]
    \draw[line width=2pt, teal!100!black] (0,0) -- (0.2,0);
\end{tikzpicture}
\small \textit{Phase 2} 

\vspace{3pt}
\textbf{Inputs: } Point clouds $\mathtt{src, tgt}$\\
\textbf{Outputs: } Transformation between point clouds $\mathtt{T}$
\begin{algorithmic}[1]
\vspace{3pt}
\State $\mathtt{iter} = \mathtt{0}$, $\mathtt{error} = \infty$
\While{$\mathtt{error} > \mathtt{tolerance}$ \textit{and} $\mathtt{iter} < \mathtt{max\_iter}$} 
    \State \quad
    \begin{tikzpicture}[overlay]
        \draw[line width=2pt, orange!80!black] (-0.5,0.25) -- (-0.5,-0.05);
    \end{tikzpicture}
    $\mathtt{adj} = \mathbf{GetCorrespondences}(\mathtt{\mathtt{src}, \mathtt{tgt}})$
    \State \quad
    \begin{tikzpicture}[overlay]
        \draw[line width=2pt, teal!100!black] (-0.5,0.2) -- (-0.5,-1.85);
    \end{tikzpicture}
    $\mathtt{T} = \mathbf{GetTransform}(\mathtt{\mathtt{src}, \mathtt{tgt},} \mathtt{adj})$
    \State \quad \hspace{0.2pt} $\mathtt{error} = \mathbf{GetError}(\mathtt{T(\mathtt{src}), \mathtt{tgt}})$
    \If{$\mathtt{error} > \mathtt{tolerance}$}
        \State \quad $\mathtt{iter} = \mathtt{iter} + 1$
        \State \quad $\mathtt{src} = \mathtt{T(\mathtt{src})}$ 
    \EndIf
\EndWhile
\State \textbf{return} $\mathtt{T}$ 
\end{algorithmic}
\end{algorithm}

\begin{figure*}[t!]
    \centering
    \includegraphics[width=0.82\textwidth]{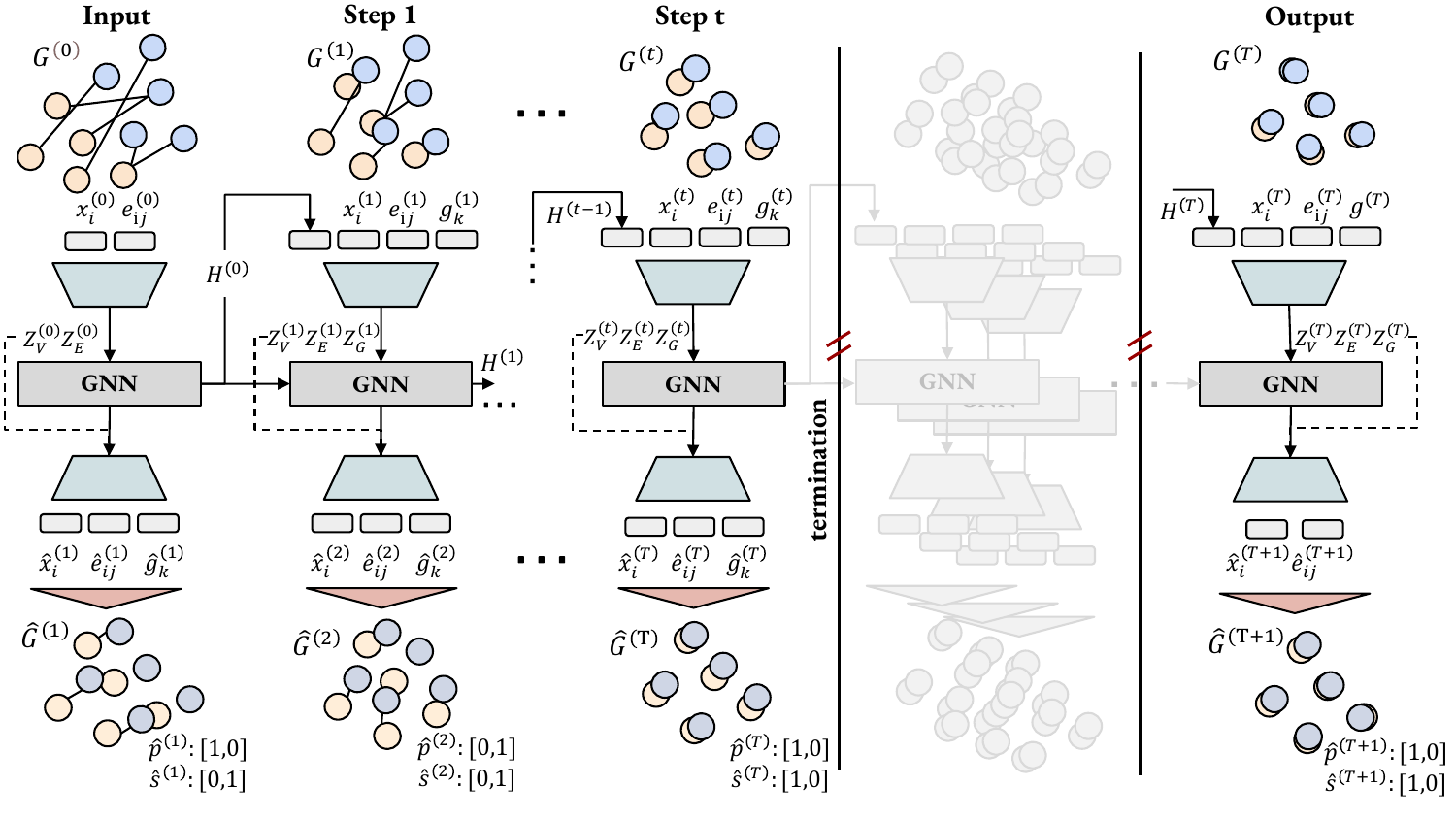}
    \caption{Learning process: Each intermediate algorithmic output is converted into a graph, $G^{(t)} = (V^{(t)}, E^{(t)}, x_i^{(t)}, e_{ij}^{(t)}, g_k^{(t)})$, before being passed to an \textit{encoder-processor-decoder} model. The encoder generates embedding representations from the input features, $Z^{(t)}$, which are then used in a Triplet \gls{mpnn} to generate latent features $H^{(t)}$. These features are passed to a decoder that predicts the features, $\hat{y}^{(t)}$, that effectively correspond to the output of the algorithm at step $t$. The process is repeated for all intermediate steps of the algorithm. The model additionally learns a phase, $\hat{p}^{(t)}$, and a termination, $\hat{s}^{(t)}$, flag, as independent binary classes, predicting the different phases of the algorithm -- finding correspondences and estimating relative transformation and error -- and its final step. During inference, when the termination flag is triggered, the neural algorithmic execution terminates at $\hat{y}^{(T)}$. Additionally, we leverage the ground truth from each input dataset as a final training signal to optimise the model's output $\hat{y}^{(T+1)}$, on the right.}
    \label{fig:relations}
    \vspace{-12pt}
\end{figure*}

\subsection{Neural Algorithmic Reasoning}

The \gls{nar} blueprint proposes to approximate a specific algorithm $\mathcal{A}$ by mimicking not only its final output but also its intermediate \textit{steps},  by using its intermediate outputs as supervisory signals. Recent methods, such as \cite{neuralexecution, clrs}, represent the steps of the algorithm as a sequence of graphs, denoted as $\mathbf{G} = \{G^{(0)} \dots G^{(T)}\}$, where $T \in \mathbb{N}$ is the final step of the algorithm, i.e. its \textit{termination}.
At each step $t \leq T$, the graph $G^{(t)}$ is described as $G^{(t)} = (V^{(t)}, E^{(t)}, x_i^{(t)}, e_{ij}^{(t)}, g_k^{(t)})$, where $V$ and $E$ are the nodes and edges, and $x_i$, $e_{ij}$, $g_k$ are the node, edge, and graph features respectively. The initial graph $G^{(0)}$ at the first algorithmic step is the \textit{input} to the network, and the last graph $G^{(T)}$ is the final \textit{output} of $\mathcal{A}$. At each step, $t$, the algorithm $\mathcal{A}$ produces results $y^{(t)}$ that are used as target outputs for a network to iteratively learn its sequential steps, i.e. its \textit{trajectory}, by learning node, edge, and graph feature representations at each step to predict the next.

\begin{figure}[]
    \centering
    \includegraphics[trim=110 80 150 130, clip,width=\columnwidth]{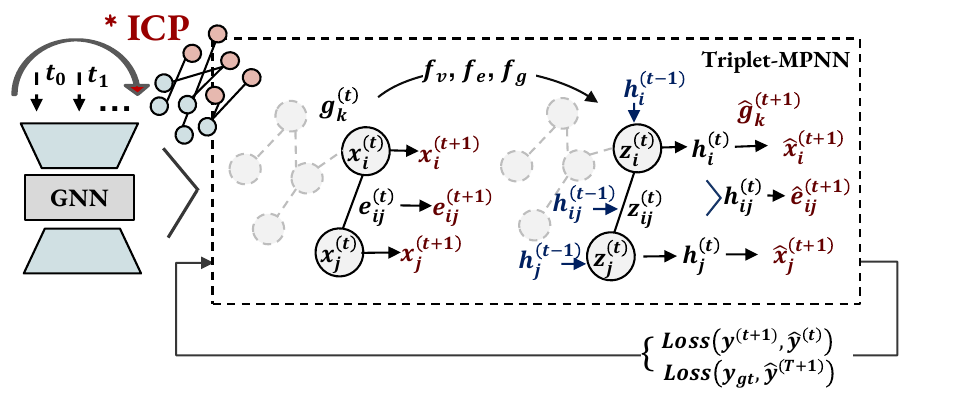}
    \caption{At each algorithmic step, we encode the input features, $x_i^{(t)}, x_j^{(t)}, e_{ij}^{(t)},$ and $g_k^{(t)}$, and use the latent features from the previous step of the processor to generate the current step's latent representations. In this context, $h_i^{(t)}$ and $h_j^{(t)}$ correspond to the node features, while $h_{ij}^{(t)}$ denotes the edge features, which result from the aggregation of node and graph encodings. The loss is calculated between the decoder's output $\hat{y}^{(t)}$ and the output of the algorithm $y^{(t)}$ at each iteration. Our method uses the ground truth from the input dataset $y_{gt}$ as an additional optimisation step $t=T+1$.}
    \label{fig:encprocdec}
    \vspace{-15pt}
\end{figure}

\subsection{NAR-*ICP Overview}

NAR-*ICP builds on the \gls{nar} blueprint and the CLRS-30 Benchmark \cite{clrs}, training \glspl{gnn} to execute classical point cloud registration algorithms using its intermediate steps as supervisory signals, effectively approximating their computational trajectory. This allows the networks to learn the inherent patterns and structure in the algorithms' reasoning process. 
We adapt the \textit{encode-process-decode} paradigm introduced in \cite{encdec} to train NAR-*ICP for our task.
\Cref{fig:relations} depicts the relation between the algorithm and the learning iterations, while \cref{fig:encprocdec} illustrates the \textit{encoder-processor-decoder} and \gls{gnn} message-passing methodology.

Ultimately, NAR-*ICP integrates geometric reasoning within the \gls{nar} framework, representing 3D and semantic point clouds from real-world datasets as dynamic bipartite graphs and learning to emulate the algorithmic trajectory through graph transformations. Unlike prior \gls{nar} approaches that assume a perfect expert algorithm, we train on both good and adversarial examples, capturing the variability and noise in real-world datasets. This improves the model’s robustness and generalisation. We further optimise the learning process using the ground truth from the dataset to refine the predictions and introduce a learned termination network that enables the model to track and predict algorithmic convergence. Our method outperforms even the algorithms it was trained on primarily because, unlike traditional \gls{icp}, it doesn't rely on a good initial alignment estimate and instead learns the initialisation. Overall, NAR-*ICP generalises well across diverse and challenging inputs, providing a robust, learning-driven alternative to classical registration algorithms and an interpretable solution to learned registration methods.

\section{Methodology}
\label{sec:methodology}

To learn to execute the intermediate steps of \gls{icp}-based algorithms, we propose a graph-based model that enables learned 3D geometric reasoning for registration, building on \gls{nar}. Specifically, we represent the input, output, and intermediate point clouds as graphs with the same structure but different features reflecting the progression of the algorithm. The input, output, and intermediate steps of the algorithm are passed to the model as \textit{probes}, i.e. the learning signals, with the intermediate steps specifically referred to as \textit{hints} \cite{clrs}, guiding the model through the reasoning process during training.

\subsection{Graph Representations}
In our model, the points in the two point clouds, $P_m$ and $P_n$, compose the nodes $V^{(t)}$ and the correspondences between them are the edges $E^{(t)}$ for each graph $G^{(t)}$, where $t$ is each step of the algorithm. Each element in $G^{(t)}$ corresponds to a probe in NAR-*ICP, as seen in \cref{tab:probes_and_features}. 


At the first step, $t=0$, we represent the initial point clouds, $P_m$ and $P_n$, as nodes $V^{(0)}$ consisting of node features, $x_{i}^{(0)}$ and $x_{j}^{(0)}$, and pass them as input probes to the model. Then, at each consecutive step $t$, we convert the output of the algorithm to a graph and pass the nodes $V^{(t)}$ and edges $E^{(t)}$, along with the graph $g_k^{(t)}$, node $x_i^{(t)}$ and $x_j^{(t)}$, and edge $e_{ij}^{(t)}$ features, as a sequence of \textit{hints}. Edge features, $e_{ij}^{(t)}$, are set to the distance in Euclidean space of the nodes $x_i^{(t)}$ and $x_j^{(t)}$. 
By representing and tracking point cloud transformations through these graph representations, we enable 3D geometric reasoning within NAR, critical for \gls{icp}-based point cloud registration. To showcase the flexibility of our method, we design different feature representations for $x_i^{(t)}$ and $x_j^{(t)}$, namely 3D spatial data, spatial data augmented with high-level semantics, and contrastively learned features, to support real-world datasets.

Additionally, we define graph features, $g_k^{(t)}$, that capture the algorithm's termination criteria: error $\mathrm{\mathbf{e}}^{(t)}$ and iterations counter. The error reflects the algorithm's convergence progress based on a pre-defined tolerance threshold, and the iterations counter contains the number of refinement steps $t$ performed to align the two point clouds. Beyond these algorithmic indicators, we also introduce two additional hint probes, phase and stop, and represent them as graph features $g_k^{(t)}$. As \gls{icp}-based algorithms are modular, the phase hint corresponds to the two main phases: (1) finding correspondences and (2) estimating the relative transformation and error between them. The stop hint guides the termination network and indicates the iteration at which the algorithm terminates.

\subsection{Probe Types}
Depending on the type of features they hold, probes are split into \textit{node}, \textit{edge}, and \textit{graph} probes to ensure data are being distributed and handled appropriately in the \gls{gnn}. Probes are categorised into \textit{scalar}, \textit{mask}, and \textit{categorical}, following the CLRS-30 Benchmark \cite{clrs}.  
Specifically, in NAR-*ICP, probes are categorised as follows:
\begin{itemize}[itemsep=0pt]
    \item \textit{Scalar} probes refer to regressed floating point variables, either in scalar or vectorial form. Here, these correspond to the point clouds, node positions\footnote{Node positions, guided by the node indices, are used in CLRS-30 to generate positional encodings for each input node.}, relative distances between correspondences, error, and iterations. 
    \item \textit{Mask} probes correspond to binary categorical features. In our method, these are the correspondences implemented as adjacency matrices.
    \item \textit{Categorical} probes are k-class categorical features. In our case, they are used for the phase and stop signals, with $k=2$ for each.
\end{itemize}

Here, we extend the CLRS-30 Benchmark to include scalar probes in the output of the \gls{nar} processes.

We normalise all input scalar probes, apart from error, using min-max scaling to ensure each feature contributes equally to the learning process and to improve convergence speed and training stability. To ensure the error probe falls within the normalisation boundaries for a maximum tolerance of $10^{-10}$, we adjust its value as follows:
\begin{equation}
\mathtt{f}_\mathrm{\mathbf{e}}(\mathrm{\mathbf{e}}) = \frac{1}{1 + e^{-\log(\mathrm{\mathbf{e}}) + c}}
\end{equation}
Here, $c=5$ is responsible for adjusting the input range for the sigmoid function and controlling the sensitivity of the changes of $\mathrm{\mathbf{e}}$.
This sigmoid transformation effectively maps the tolerance into a range of (0,1) with the behaviour controlled by the magnitude of $\mathrm{\mathbf{e}}$.

\begin{table}[]
    \centering
    \small
    \caption{Input, hints, and output probes, probe types, and associated graph features in NAR-*ICP.}
    \renewcommand{\arraystretch}{1.4}
        \setlength{\tabcolsep}{5pt}

    \begin{tabular}{c||c|c|c}
         & Probe & Feature & Type \\
        \hline \hline
        \multirow{2}{*}{\rotatebox[origin=c]{90}{\centering Input}} 
        & Point clouds & $x_i^0, x_j^0$ & scalar  \\
        & Node positions & $V^0$ & scalar\\
        \hline
        \multirow{8}{*}{\rotatebox[origin=c]{90}{\centering Hints}} 
        & Transformed Point clouds & $x_i^{(t)}, x_j^{(t)}$ & scalar \\
        & Correspondences & $E^{(t)}$ & mask \\
        & Distances & $e_{ij}^{(t)}$ & scalar \\
        & Error & $g_k^{(t)}$ & scalar \\
        & Iterations & $g_k^{(t)}$ & scalar \\
        & Phase & $g_k^{(t)}$ & categorical \\
        & Stop & $g_k^{(t)}$ & categorical \\
        \hline
        \multirow{2}{*}{\rotatebox[origin=c]{90}{\centering Output}} 
        & Final Point clouds & $x_i^{(T)}, x_j^{(T)}$ & scalar \\
        & Final Correspondences & $E^T$ & mask \\
    \end{tabular}
    \label{tab:probes_and_features}
    \vspace{-13pt}
\end{table}

\subsection{Hint Updates}\label{subsec:hintupdates}
All key algorithmic computations are encapsulated in the hint probes used as training signals for NAR-*ICP. Hints capture the trajectory of the algorithms and thus reflect the different aspects of their internal computations. We design our method to support flexible hint configurations, able to adapt to different algorithmic variants, and integrate diverse data modalities from real-world datasets, including semantics, surface normals, and covariances. While ICP-based algorithms are structurally similar, as seen in \Cref{algo:icp}, the specific intermediate outputs and calculations, such as error and relative transformation computations, generate distinct hint trajectories for each variant. 

To effectively train NAR-*ICP to mimic the behaviour of the algorithms, we aim to capture as many algorithmic details as possible in the hint update configuration. We provide a pseudocode, in \Cref{algo:nar}, illustrating the modifications made to \Cref{algo:icp} in order to perform the appropriate hint updates for NAR-*ICP. In  \Cref{algo:nar}, node hints are highlighted in $\textcolor{myred}{\mathtt{blue}}$, edge hints in $\textcolor{mygreen}{\mathtt{red}}$, and graph hints in $\textcolor{myblue}{\mathtt{green}}$. To differentiate between the stages of the reasoning process, we use the following notation: input probes are denoted without a subscript, intermediate hints are marked with subscript $*\mathtt{_h}$, and final probes use the subscript $*\mathtt{_{final}}$.

\textbf{Multi-phase approach. }In particular, to capture the behaviour and intermediate computations of the two-phase ICP-based algorithms in this study, we update the hints after each phase. Here, \textit{Phase 1} denotes the correspondence-finding phase, in $\mathbf{GetCorrespondences}$, and \textit{Phase 2} the relative transformation and error estimation phase, in $\mathbf{GetTransform}$ and $\mathbf{GetError}$.
The second phase is different for each algorithm.
During inference, the $\textcolor{myblue}{\mathtt{phase}}$ hint predictions allow us to identify the intermediate components of the neural algorithmic execution, while the $\textcolor{myblue}{\mathtt{stop}}$ hint predictions determine the termination of the processor's iterations.

\textbf{Parallelised hint design. }In ICP-based algorithms, there are multiple control flow statements that can generate different algorithmic trajectories. As \glspl{gnn} operate on parallelised architectures, we ``compressed'' the hints to align with the network's training behaviour while still capturing succinct yet descriptive algorithmic trajectories. For instance, in the correspondence-finding phase, each point in $\textcolor{myred}{\mathtt{tgt_h}}$ is iteratively compared to the points in $\textcolor{myred}{\mathtt{src_h}}$ to find its nearest neighbour. Following the parallelised nature of the underlying network, we optimise this process by passing only the resulting correspondences as an adjacency matrix, $\textcolor{mygreen}{\mathtt{adj_h}}$, and the distance matrix of the correspondences, $\textcolor{mygreen}{\mathtt{dist}}$, within the hints trajectory. Similarly, after estimating the relative transformation between the two point clouds, we transform $\textcolor{myred}{\mathtt{src_h}}$ and update the hint.

\textbf{Termination criteria. }The intermediate algorithmic steps, denoted as $t$, that guide the iterative training process of the network, effectively capture the evolving trajectory of the algorithm as it progresses through each phase on every iteration. This process continues until the algorithmic termination criteria are met. In particular, the algorithm terminates if the $\textcolor{myblue}{\mathtt{error}}$ drops below the $\mathtt{tolerance}$ threshold or if $\textcolor{myblue}{\mathtt{iter}}$ reaches the maximum number of iterations in $\mathtt{T_{max}}$.

\textbf{Optimisation. }In NAR-*ICP, we introduce an additional optimisation step using the ground truth obtained from the input datasets. In the $\mathbf{GetGT}$ function, we use the ground-truth relative transformation to find the ground-truth correspondences. To do so, we transform the initial $\mathtt{src}$ point cloud using the ground-truth relative transformation to align the two point clouds. Then, for each point in the transformed $\mathtt{src}$, we find its closest points in $\mathtt{tgt}$ and extract the ground truth correspondences. If the $\mathtt{gt\_optimisation}$ flag is triggered, we pass the transformed $\mathtt{src}$, the $\mathtt{tgt}$, and the ground truth correspondences as output probes $\mathtt{src_{final}}$, $\mathtt{tgt_{final}}$, and $\mathtt{adj_{final}}$, respectively. If the flag is not triggered, we pass the final output of the algorithm as the output probe.

\begin{algorithm}[t]
\caption{Probes and hint updates$^\texttt{*}$ for generating the supervisory signals for NAR-*ICP. \label{algo:nar}}
\begin{tikzpicture}[baseline]
    \draw[line width=2pt, orange!80!black] (0,0) -- (0.2,0);
\end{tikzpicture}
\small \textit{Phase 1}
\begin{tikzpicture}[baseline]
    \draw[line width=2pt, teal!100!black] (0,0) -- (0.2,0);
\end{tikzpicture}
\small \textit{Phase 2}
\quad \quad \quad \quad \quad \quad \quad \small $^\texttt{*}$\textit{denoted by $\leftarrow$ assignment}
\vspace{3pt}

\textbf{Input: }Point clouds $\mathtt{src, tgt}$. Node positions $\mathtt{pos}$. \\
\textbf{Hints: }Point clouds $\textcolor{myred}{\mathtt{src_h}}, \textcolor{myred}{\mathtt{tgt_h}}$. Correspondences $\textcolor{mygreen}{\mathtt{adj_h}}$. Distances $\textcolor{mygreen}{\mathtt{dist}}$. Iterator $\textcolor{myblue}{\mathtt{iter}}$. Error $\textcolor{myblue}{\mathtt{error}}$. Phase $\textcolor{myblue}{\mathtt{phase}}$. Termination $\textcolor{myblue}{\mathtt{stop}}$. \\
\textbf{Output: }Final point clouds ${\mathtt{src_{final}, tgt_{final}}}$. Final correspondences ${\mathtt{adj_{final}}}$

\begin{algorithmic}[1]

\vspace{4pt}
\State $\textcolor{myred}{\mathtt{src_h}} \leftarrow  \mathtt{src} , \textcolor{myred}{\mathtt{tgt_h}} \leftarrow  \mathtt{tgt}$
\State $ \textcolor{myblue}{\mathtt{stop}} \leftarrow \mathtt{False}, \textcolor{myblue}{\mathtt{error}} \leftarrow \infty, \textcolor{myblue}{\mathtt{iter}} \leftarrow 0$
\While{$\textcolor{myblue}{\mathtt{error}} > \mathtt{tolerance}$ \textit{and} $\textcolor{myblue}{\mathtt{iter}} < \mathtt{T_{max}}$} 
\State \quad
    \begin{tikzpicture}[overlay]
        \draw[line width=2pt, orange!80!black] (-0.5,0.2) -- (-0.5,-0.78);
    \end{tikzpicture}
     \hspace{-5.1pt} $\textcolor{myblue}{\mathtt{phase}} \leftarrow 0$
    \State \quad $\textcolor{mygreen}{\mathtt{adj_h}} \leftarrow \mathbf{GetCorrespondences}(\textcolor{myred}{\mathtt{src_h}}, \textcolor{myred}{\mathtt{tgt_h}})$
    \State \quad $\textcolor{mygreen}{\mathtt{dist}} \leftarrow \|(\textcolor{myred}{\mathtt{src_h}} - \textcolor{myred}{\mathtt{tgt_h}})\|_{\textcolor{mygreen}{\mathtt{adj_h}}}$
\State \quad 
    \begin{tikzpicture}[overlay]
        \draw[line width=2pt, teal!100!black] (-0.5,0.2) -- (-0.5,-3);
    \end{tikzpicture}
     \hspace{-5.1pt} $\textcolor{myblue}{\mathtt{phase}} \leftarrow 1$
    \State \quad $\mathtt{T} = \mathbf{GetTransform}(\textcolor{myred}{\mathtt{src_h}}, \textcolor{myred}{\mathtt{tgt_h}})_{\textcolor{mygreen}{\mathtt{adj_h}}}$
    \State \quad $\textcolor{myblue}{\mathtt{error}} \leftarrow \mathbf{GetError}(\mathtt{T(\textcolor{myred}{\mathtt{src_h}}), \textcolor{myred}{\mathtt{tgt_h}}})$
    \If{$\textcolor{myblue}{\mathtt{error}} > \mathtt{tolerance}$ \textit{and} $\textcolor{myblue}{\mathtt{iter+1}} < \mathtt{T_{max}}$}
        
        \State \quad $\textcolor{mygreen}{\mathtt{dist}} \leftarrow \|(\mathtt{T(\textcolor{myred}{\mathtt{src_h}})-\textcolor{myred}{\mathtt{tgt_h}}})\|_{\textcolor{mygreen}{\mathtt{adj_h}}}$
        \State \quad $\textcolor{myred}{\mathtt{src_h}} \leftarrow \mathtt{T(\textcolor{myred}{\mathtt{src_h}})}$
        \State \quad $\textcolor{myblue}{\mathtt{iter}} \leftarrow \textcolor{myblue}{\mathtt{iter}} + 1$
    \Else
        \State \quad $\textcolor{myblue}{\mathtt{stop}} \leftarrow \mathtt{True}$
        \State \quad $\mathtt{\textcolor{black}{\mathtt{adj_{final}}, \mathtt{src_{final}}} = \textcolor{mygreen}{\mathtt{adj_h}}, \mathtt{T(\textcolor{myred}{\mathtt{src_h}}}})$
    \EndIf
\EndWhile
\If{$\mathtt{gt\_optimisation}$}
    \State \quad $\mathtt{T}, \textcolor{black}{\mathtt{adj_{final}}} = \mathbf{GetGT}(\mathtt{src, tgt})$
    \State \quad $\mathtt{src_{final} = T(src)}$
\EndIf
\State $\mathtt{tgt_{final} = tgt}$
\State \textbf{return} $\textcolor{black}{\mathtt{{src_{final}, tgt_{final}}}}, \textcolor{black}{\mathtt{adj_{final}}}$
\end{algorithmic}

\end{algorithm}

\subsection{Model Architecture }
The \gls{gnn} architecture builds on the encode–process–decode paradigm \cite{encdec}, adapted and extended to effectively approximate complex registration algorithms through 3D geometric reasoning and handle real-world datasets.

\textbf{Encoders. }To process the input features, we define linear \textit{encoders} $\mathtt{f_v}$, $\mathtt{f_e}$, and $\mathtt{f_g}$ for each $x_i$, $e_{ij}$, $g_k$ to generate their corresponding embedded representations at the node, edge, and graph level, respectively:
\begin{equation}
\begin{aligned}
Z_V^{(t)} &= \{ z_i^{(t)} = \mathtt{f_v}(x_i^{(t)}) \mid \forall x_i^{(t)} \in V^{(t)} \} \\
Z_E^{(t)} &= \{ z_{ij}^{(t)} = \mathtt{f_e}(e_{ij}^{(t)}) \mid \forall e_{ij}^{(t)} \in E^{(t)} \} \\
Z_G^{(t)} &= \{ z_k^{(t)} = \mathtt{f_g}(g_k^{(t)}) \mid \forall g_k^{(t)} \in G^{(t)} \}
\end{aligned}
\end{equation}
We call the collection of such embeddings  $Z^{(t)}$.

Here, we note that although the target point cloud $P_n$ remains unchanged during registration, it is passed as a hint at each iteration, and gets re-encoded to guide the network in reasoning over correspondences as the source point cloud $P_m$ changes. Effectively, with this design, the network learns that $P_n$ should remain fixed throughout, which is fundamental to \gls{icp}'s iterative reasoning process.

Additionally, the second hint ($t=1$) corresponds to the initial alignment estimate used in the initialisation step of \gls{icp} algorithms, and serves as a training signal for NAR-*ICP. This enables the model to learn the initialisation, instead of relying on it during inference, addressing a key limitation of traditional \gls{icp} methods.

\textbf{Processor. }The \textit{processor} network is a recurrent \gls{gnn} and is responsible for the major part of the learning process. The \gls{gnn}'s message-passing methodology is depicted in \cref{fig:encprocdec}. It takes as input the embedding $Z^{(t)}$ from the current step, as well as the previous latent features ${H}^{(t-1)}$ to compute the latent features from the current step using \gls{mpnn} \cite{mpnn, clrs}: 
\begin{equation}
    H^{(t)} = \mathtt{P}(Z^{(t)}, H^{(t-1)})
\end{equation}
Where for each node:
\begin{equation}
\begin{aligned}
    h_i^{(t)} &= \mathtt{f_r}(z_i^{(t)}\|h_i^{(t-1)}, m_i^{(t)}) \\
    m_i^{(t)} &= \max_{1 \leq j \leq n} \mathtt{f_m} \left( z_i^{(t)}\|h_i^{(t-1)}, z_j^{(t)}\|h_j^{(t-1)}, z_{ij}^{(t)}, z_k^{(t)} \right)
\end{aligned}
\end{equation}
Here, $\mathtt{f_m}$ is the message-passing function of the \gls{mpnn}, $\mathtt{f_r}$ is the readout function, and the embedding calculations start with $ h_i, h_{ij} = \underline{0} ~ \forall  h_i, h_{ij} \in H^{(t)}$ in the input step $t=0$ \cite{generalistNAR}. As our task requires both edge and node-level reasoning, message-passing needs to also be performed in the edge embeddings \cite{dynamicprogrammers}, through \textit{triplet reasoning} \cite{generalistNAR}. Following the architecture of Triplet-\gls{mpnn}, we compute representations over node-edge-node triplets, to obtain edge embeddings:
\begin{equation}
t_{ij}=\mathtt{t_e}(h_i,h_j,z_{ij}^{(t)},z_g^{(t)}) \quad \text{and} \quad  h_{ij}=\mathtt{t_r}(\max t_{ij})
\end{equation}
Where $\mathtt{t_e}$ is a function that generates the triplet embeddings and $\mathtt{t_r}$ is a readout function used to extract each respective edge embedding.


\textbf{Decoders. }The \textit{decoders}, $\mathtt{g_v}$, $\mathtt{g_e}$, and $\mathtt{g_g}$, then decode the outputs for each step as:
\begin{equation}
\begin{aligned}
\hat{x}_i^{(t+1)} &= \mathtt{g_v}(\hat{z}_i^{(t)}, h_i^{(t)}), \\ 
\hat{e}_{ij}^{(t+1)} &= \mathtt{g_e}(\hat{z}_{ij}^{(t)}, h_{ij}^{(t)}), \\  \hat{g}_k^{(t+1)} &= \mathtt{g_g}(\hat{z}_{k}^{(t)}) 
\end{aligned}
\end{equation}
We call the collection of these outputs $\hat{y}^{(t)}$.
Each decoded output effectively corresponds to the predicted input graph representation of the next step.

\textbf{Termination. } 
In NAR-*ICP, we introduce learned termination that predicts when to stop the iterations, effectively tracking algorithmic convergence. A dedicated termination network is implemented as a binary classification task. The network is trained in the processor using a separate mask probe through the $\textcolor{myblue}{\mathtt{stop}}$ hint and gets encoded as a graph probe alongside other termination criteria, i.e. $\textcolor{myblue}{\mathtt{error}}$, $\textcolor{myblue}{\mathtt{iter}}$, and $\textcolor{myblue}{\mathtt{phase}}$. Using the network's predictions, the processor dynamically adjusts the number of iterations required during inference, thus avoiding unnecessary computations and improving the overall efficiency and convergence of the iterative process. In particular, the termination network predicts the probability of each iteration being the final step of the algorithm and, consequently, the termination step of the processor, effectively halting further iterations.

\subsection{Training}
Each probe type is handled differently during training, both in the \textit{encoder-processor-decoder} and the loss calculations, depending on whether it belongs to node, edge, or graph probes, and its specific data type, i.e. scalar, mask, or categorical. 

In the \textit{encoder-processor-decoder}, node features are initially encoded individually, updated through message-passing between node, edge, and graph features, and then decoded into node-specific outputs. Edge features are encoded, capturing relationships between node pairs, and then get iteratively updated, forming triplets of node-edge-node pairs before being decoded. Graph features are encoded and updated globally, aggregating information from across the graph, and finally, get decoded into global outputs that represent overall graph properties. This process is further described in \Cref{sec:overview}.

In the loss calculation, the scalar, mask, and categorical features define the type of loss to be applied to each type of probe. In particular, for each hint and output probe, we compare the values extracted from the decoders' predictions $\hat{y}^{(t)}$ with the ground truth hints $y^{(t)}$ at each processor step $t$ and calculate the appropriate loss depending on its type. The losses used in NAR-*ICP are as follows:
\begin{itemize}
    \item For the \textit{scalar} probes, we apply a \gls{mse} loss with a scaling factor $\alpha$ to balance the loss terms and stabilise the training:
    \begin{equation}
        \mathcal{L}_{scalar}^{(t)} = \alpha \frac{1}{|Y_s|} \sum_{y_i \in Y_s} (y_i^{(t)} - \hat{y}_i^{(t)})^2
    \end{equation}
    where $Y_s$ is the set of scalar variables in $y$.
    \item For the \textit{mask} probes, we introduce the set of mask variables $Y_m$, in $y$ and apply a weighted \gls{bce} loss:
    \begin{align}
     w_{pos}^{(t)} &= \frac{\sum_{y_i \in Y_m} (1 - y_i^{(t)})}{\sum_{y_i \in Y_m} y_i^{(t)} + \epsilon} \\
    \begin{split}
    \mathcal{L}_{mask}^{(t)} &=  -\frac{1}{|Y_m|} \sum_{y_i \in Y_m} \left( w_{pos}  y_i^{(t)} \log(\hat{y}_i^{(t)}) \right. \\& \left. + (1 - y_i^{(t)}) \log(1 - \hat{y}_i^{(t)}) \right)
    \end{split}
    \end{align}
    \item For the \textit{categorical} probes, we apply a \gls{cce} loss with $Y_c$ the set of scalar variables in $y$:
    \begin{equation}
    \mathcal{L}_{categ}^{(t)} = -\frac{1}{|Y_c|}\sum_{y_i \in Y_c} \sum_{c \in k} y_{i,c}^{(t)} \log(\hat{y}_{i,c}^{(t)})
    \end{equation}
\end{itemize}

We also introduce an optimisation step to our training process, as described in \Cref{subsec:hintupdates}. This includes passing the ground truth transformed point clouds and correspondences as additional training signals in the final step of the model and computing an additional loss term, as shown in the last lines of \Cref{algo:nar}. We retrieve the ground truth information from the input datasets. The last step of our model is then trained conditionally:
\begin{equation}
    \mathcal{L}_{final} = 
\begin{cases} 
\mathcal{L}(\hat{y}^{(t)}, y_{gt}), & t=T+1 \\
\mathcal{L}(\hat{y}^{(t)}, y^{(t)}), & t=T
\end{cases}
\end{equation}
Here, $\mathcal{L}$ corresponds to the $\mathcal{L}_{scalar}$ loss for the point clouds and the $\mathcal{L}_{mask}$ loss for the correspondences. We retrieve the ground truth transformed source point clouds as $P_{gt} = R_{gt}P_m + t_{gt}$, and then find the closest points between $P_n$ and $P_{gt}$ to extract the ground truth adjacency matrix, $adj_{gt}$. Unlike traditional \gls{nar}, which assumes a perfect expert, our approach relaxes this assumption by training on both adversarial and good examples, and directly optimising against the ground truth from the dataset.

To ensure input hints are encoded in each step of the processor, we use \textit{teacher-forcing} \cite{teacher} where we feed back the ground-truth hints in the input during training to stabilise the trajectory of the hints. Naturally, this is done only during training as, during inference, the encoded hints are equal to the decoded hints of the previous step. Due to the recursive nature of \gls{nar} and the large number of iterations when training NAR-*ICP, the issues of exploding and vanishing gradients are prominent. To tackle these, we apply Xavier initialisation \cite{xavier} on the scalar hints and gradient clipping \cite{gradientclipping}. We also apply layer normalisation in the iterative model to stabilise the learning process further and enhance generalisation.


\section{Experimental Setup}
\label{sec:experiments}

In this section, we discuss our experimental setup, detailing the specific algorithms, baselines, datasets, tools, and configurations used to evaluate our approach.

\subsection{Algorithms}
As our approach is generic for point cloud registration, we evaluate it in approximating several established algorithms from the \gls{icp} family: point-to-point ICP (\texttt{P2P-ICP}) \cite{icp}, point-to-plane ICP (\texttt{P2L-ICP}) \cite{point-plane}, and Generalised-ICP (\texttt{G-ICP}) \cite{gicp}.
These algorithms also serve as the non-learned baselines for evaluating our method, further assessing its generalisation capabilities.

While all calculate correspondences $C^{(t)}$ at each iteration as the closest points in the transformed $P_m$ for each point in $P_n$ in Euclidean space, they differ in the calculation of the relative transformation $[\mathbf{R}^{(t)}_{m,n} \mid \mathbf{t}^{(t)}_{m,n}]$ and the error $\mathrm{\mathbf{e}}^{(t)}$. The iterative process repeats until convergence, i.e. the transformation estimations stabilise and the error is minimised. 
\Cref{algo:icp} gives an overview of the generic \gls{icp} operations.

\begin{enumerate}[wide, labelindent=10pt]
\item \textbf{Point-to-point ICP. } 
After finding correspondences $C^{(t)}$, \texttt{P2P-ICP} estimates the relative transformation between them, $[\mathbf{R}^{(t)}_{m,n} \mid \mathbf{t}^{(t)}_{m,n}]$, through \gls{svd}. The transformation is then used to transform $P_m$, and the process repeats until it minimises the sum of square distances between corresponding points, denoted as the error:
\begin{equation}
\mathrm{\mathbf{e}}^{(t)} = \sum_{(\mathbf{p}_i, \mathbf{p}_j) \in C^{(t)}} ||\Delta_{ij}^{(t)}||_2^2
\end{equation}
We define $\Delta_{ij}^{(t)}$ as the vectorial distance between each set of points $(\mathbf{p}_i, \mathbf{p}_j) \in C^{(t)}$, as:
\begin{equation}
    \Delta_{ij}^{(t)} = \mathbf{R}^{(t)}_{m,n} \mathbf{p}_i + \mathbf{t}^{(t)}_{m,n} - \mathbf{p}_j
\end{equation}


\item \textbf{Point-to-plane ICP. } 
At each iteration, \texttt{P2L-ICP} minimises the distance between each point in $P_n$ to the tangent plane of the corresponding transformed point in $P_m$, represented by its normal vector $\mathbf{n}_j$.
The error $\mathrm{\mathbf{e}}^{(t)}$ is then computed as follows:
\begin{equation}
    \mathrm{\mathbf{e}}^{(t)} = \sum_{(\mathbf{p}_i, \mathbf{p}_j) \in C^{(t)}} (\mathbf{n}_j^\top \cdot \Delta_{ij}^{(t)})^2
\end{equation}
where $\cdot$ denotes the dot product.
The algorithm minimises the point-to-plane error, $\mathrm{\mathbf{e}}^{(t)}$, by iteratively calculating the relative transformation and solving a linear system derived by the Jacobian $J$ of the error $\mathrm{\mathbf{e}}^{(t)}$ until convergence.

\item \textbf{Generalised-ICP. }
Relying on the covariance matrices of the correspondences, \texttt{G-ICP} minimises the Mahalanobis distance between corresponding points. The algorithm combines \texttt{P2P-ICP} and \texttt{P2L-ICP} into a single optimisation framework.
We denote with $\mathbf{M}_{ij}^{(t)}$ the inverse covariance matrix between $(\mathbf{p}_i, \mathbf{p}_j) \in C^{(t)}$:
\begin{equation}
    \mathbf{M}_{ij}^{(t)} = (\mathbf{\Sigma}_{p_j}^{(t)} + \mathbf{R}^{(t)}_{m,n} \mathbf{\Sigma}_{p_i}^{(t)} {\mathbf{R}_{m,n}^{(t)\top}})^{-1}
\end{equation}
where $\mathbf{\Sigma}_{p_i}^{(t)}$ and $\mathbf{\Sigma}_{p_j}^{(t)}$ are the covariance matrices of points in $P_m$ and $P_n$, respectively.
The error function $\mathrm{\mathbf{e}}^{(t)}$ is then defined as:
\begin{equation}
    \mathrm{\mathbf{e}}^{(t)} = \sum_{(\mathbf{p}_i, \mathbf{p}_j) \in C^{(t)}} {\Delta_{ij}^{(t)\top}} \mathbf{M}^{(t)}_{ij} \Delta_{ij}^{(t)}
\end{equation}
The gradient of $\mathrm{\mathbf{e}}^{(t)}$ is followed until convergence.
\end{enumerate}

\subsection{Learned Baselines}\label{subsec:learnedbaselines}
To further validate our method as a strong registration benchmark, we compare it against \gls{icp}-based learned registration methods and adapted versions of the latest state-of-the-art.
While learned registration benchmarks typically rely on dense, large-scale point clouds, our framework leverages object-centric, low-cost point clouds that align with its algorithmic structure. As the original implementations were not directly transferable to our setting, we adapt existing baselines, where possible, and design new ones tailored to our use case. To ensure fair and meaningful comparisons, we retain the core architectural principles of each baseline while optimising them for performance, accuracy, and speed. The details of these adaptations are provided below.

\begin{enumerate}[wide, labelindent=10pt, itemsep=0.4em]
    \item \textbf{\texttt{GeoTransformer}}: GeoTransformer \cite{geotransformer} relies heavily on the geometric structure of point clouds and therefore requires the most substantial adaptation for our use case. TMP \cite{tmp}, introduces an object-centric registration framework based on GeoTransformer, but as its implementation is not publicly available, we draw inspiration from its design.
    In particular, we replace the hierarchical, coarse-to-fine registration strategy in GeoTransformer, with a direct point-to-point approach. For our baseline, we use all input points as superpoints, retaining the same geometric embedding principles and transformer-based architecture. The KPConv-FPN encoder is replaced with a lightweight 3-layer linear encoder (with layer norm, ReLU, and dropout), similar to NAR-*ICP. To enhance the model's capacity for complex feature learning, we deepened the transformer from $4$ to $8$ blocks, doubled the model dimensions (backbone: from $32/128$ to $64/256$; transformer: from $64/128$ to $128/256$), and increased the number of attention heads from $4$ to $8$. Each input point cloud is normalised independently, and correspondences are found using geometric proximity and a simplified optimal transport fallback. We tune the geometric embedding parameters, setting the distance and angular embedding temperatures to $3$ and $8$, respectively, and increasing the angular neighbours to $5$. Training is stabilised by lowering the learning rate to $10^{-6}$, increasing weight decay to $10^{-4}$, adding gradient clipping ($norm = 0.5$), and extending to 300 epochs. We also enhance SVD stability in weighted procrustes by adding regularisation to the input matrix before decomposition, validate rotation matrices, and add robust fallbacks for sparse matches. 
    \item \textbf{\texttt{Predator}}: We modify Predator \cite{predator} to support sparse point clouds, while preserving its KPFCNN and \glspl{gnn}-based architecture. The model parameters were tuned for small-scale inputs by setting input and \gls{gnn} features' dimensions to $256$, reducing final feature dimensions to $32$, and applying finer subsampling to address sparsity. The loss function is restructured to replace radius-based filtering with a closest-match approach in both circle and saliency losses. Training is further stabilised with a learning rate of $0.01$, improved error handling, and normalisation of input point clouds. The data pipeline is adapted for dynamic point cloud sizes, using a resampling strategy that retains all points when input sizes fall below the target. Additionally, the adapted model supports minimal-overlap correspondence scenarios found in small-scale point cloud data. 
    \item \textbf{\texttt{DiffICP}}: We introduce a differentiable variant of the traditional \gls{icp} algorithm. DiffICP predicts the final transformation in a single forward pass, instead of operating iteratively like classical ICP and NAR-*ICP. The architecture consists of several key components: a registration network that uses multiple \glspl{mlp} of size $64$ and $128$ to extract features from normalised input point clouds, followed by a differentiable correspondence selection mechanism using softmax-based similarity matching, and finally, a differentiable SVD-based transformation estimation. The model is trained using a dual loss training strategy combining \gls{mse} loss on the predicted transformations with \gls{bce} loss on the correspondences. The model is optimised using Adam optimiser with a learning rate of $10^{-3}$ and weight decay of $10^{-4}$.
    \item \textbf{\texttt{DGR}}: In DeepGlobalRegistration (DGR) \cite{dgr}, we retain the original FCGF \cite{fcgf} feature extraction pipeline but adapt the registration pipeline. Voxel downsampling has been removed and point cloud normalisation has been introduced. We enhance the SVD-based Procrustes module with adaptive regularisation and dynamic weight tensor handling for numerical stability. The correspondence detector now supports sparse inputs by reducing the required number of correspondences from $200$ to as few as $10$, or $10\%$ of the total points. The inlier prediction network has been modified to produce fallback confidence outputs when few correspondences are detected. We also adjusted the training configuration, reducing neighbour limits to $50$ and tuning weight clipping thresholds. The safeguard registration mechanism has been modified to effectively disable radius filtering, ensuring all available correspondences are utilised in the registration process.
    \item \textbf{\texttt{DCP}}: Deep Closest Point \cite{dcp}, has been restructured to support sparse point clouds through dynamic graph construction. Hard correspondence assignments are replaced with soft, probabilistic mappings in the SVD module, where each source point receives a weighted combination of target points based on feature similarity rather than strict one-to-one mappings. The k-nearest neighbour computation is made input-aware, removing reliance on fixed thresholds.  We extract embeddings from dynamically padded, normalised input point clouds using DGCNN \cite{dgcnn} and a Transformer \cite{attention_is_all_you_need}. The training process was restructured to remove cycle consistency, which introduced unnecessary complexity. Instead, we implemented circular prediction, where the model is applied iteratively ($5$ iterations) to progressively transformed source point clouds. The predicted transformations are accumulated at each step, allowing the system to refine the alignment incrementally.
\end{enumerate}

\subsection{NAR in CLRS-30}\label{subsec:clrs30nar}

In addition to the main baselines, we also evaluate our methodology against certain architectural features from the original NAR implementation in the CLR-30 Benchmark, such as the number of processor steps and the final output signal, by reproducing them within our NAR-*ICP framework. The original NAR implementation is not directly compatible or comparable with our framework, as it is developed and optimised for the algorithms in CLRS \cite{clrs_book} and synthetic datasets. By transferring its key features to our setting, we implicitly compare against NAR in CLRS-30, isolating and assessing the contribution of our architectural optimisations in a directly comparable setting. We are particularly interested in the number of processor steps, as it is a known limitation of the CLRS-30 Benchmark. We refer to this baseline as \texttt{NAR-CLRS}. When specifically evaluating the number of processor steps, we denote it as NAR-*ICP (last) in our result tables to facilitate comparisons and visualisation.

\subsection{Datasets}

To thoroughly evaluate our system, we consider two distinct datasets: one synthetic and one real-world, each with very different characteristics.

\textbf{Synthetic.} We build a complex synthetic dataset consisting of random point clouds with $(x, y, z)$ coordinates uniformly distributed within the range $[-40, 40]$. Each point cloud is part of a set where the second point cloud is generated by applying a random relative transformation to the first. This dataset is particularly interesting because ICP-based algorithms are expected to underperform or even fail to converge on point clouds with large relative transformations.
    
\textbf{Real-World.} Our real-world dataset consists of centroids of objects and associated semantic labels extracted from SemanticKITTI \cite{semantickitti}. Here, we define as \textit{object} a local region in the scan with the same semantic class. To extract them, we cluster points within each semantic class based on spatial proximity, with constraints on minimum cluster size and maximum clustering tolerance as in \cite{Kong2020sgpr,semgat}.
The centroids are then computed as the mean of the coordinates of all points belonging to an object. To assess the performance of our method, we evaluate our approach on scan pairs recorded \SI{1.6}{\metre}, \SI{11.3}{\metre}, and \SI{24}{\metre} apart on average. In registration tasks, scans recorded further apart have lower overlap and are more challenging than scans recorded closer together due to the larger number of outliers and correspondence mismatches.

We demonstrate the flexibility of our method in handling various types of inputs, as we solve the registration either using 3D coordinates alone for each centroid or by additionally including the associated semantic class. For the real-world datasets, we run the algorithms using point cloud coordinates, but train the models also using semantics passed as node probes and encoded along with the coordinates.



\subsection{Metrics}
\label{sec:metrics}
We employ several quantitative metrics to evaluate the performance of our proposed approach compared to the baselines. These metrics are applied to different combinations of outputs and target signals, as outlined below:
\begin{enumerate}[itemsep=0pt]
    \item \texttt{*$^{\textbf{T}}$}: The final step of the neural algorithmic execution, $\hat{y}^{(T)}$, as predicted from the termination, compared to the final step of the algorithm, $y^{(T)}$.
    \item\texttt{*$^{\textbf{t}}$}: Each step of the neural algorithmic execution, $\hat{y}^{(t)}$, compared to each step of the algorithm, $y^{(t)}$.
    \item \texttt{*$^{\textbf{GT}}$}: The final step of the algorithm, $y^{(T)}$, and the final step of the neural algorithmic execution -- either before, as predicted from the termination, $\hat{y}^{(T)}$, or after training with ground truth optimisation, $\hat{y}^{(T+1)}$ -- compared to the ground truth from the input dataset, $y_{gt}$.
\end{enumerate}
We utilise the definition above and apply it to our metrics, each denoted with the corresponding superscript, e.g. \texttt{[metric]}$^{\textbf{T}}$.

\textbf{Registration Accuracy. }
First, we assess the performance of our method as a pose regressor, which is the primary goal of point cloud registration algorithms. We use the \gls{rte}$(\downarrow)$ and \gls{rre}$(\downarrow)$ from the KITTI metrics \cite{kittimetrics}, where:
\begin{equation}
\begin{aligned}
RTE(t) &= \|t - \hat{t}\| \\
RRE(R) &= \arccos \frac{\text{Tr}(\hat{R}^\top R) - 1}{2}
\label{eq:rte_rre}
\end{aligned}
\end{equation}
Here, $\hat{t}$ and $\hat{R}$ are the estimated translation and rotation from the model, $R$, $t$ are the ground truth values, and $Tr(\cdot)$ represents the trace of a matrix. Lower values in \gls{rte} and \gls{rre} metrics \cite{kittimetrics} indicate better performance. We compare our approach against the original algorithm used for its training, in RTE$^{\textbf{T}}$ and RRE$^{\textbf{T}}$, and the ground truth, in RTE$^{\textbf{GT}}$ and RRE$^{\textbf{GT}}$. To retrieve the estimated relative transformations from NAR-*ICP, we apply \gls{svd} on the nodes corresponding to the predicted target $P_n$ and transformed source $\hat{P}_m$ point clouds.

\textbf{Algorithmic Alignment.}
To evaluate the accuracy of the intermediate hints predictions, we extract the \gls{mse}$^{\textbf{t}}$ error distributions of the predicted transformed point clouds against those of the algorithm at each step of the neural algorithmic execution. We then compare their medians, interquartile range (IQR), and number of outliers for each benchmark and dataset. Lower medians and smaller IQRs indicate more accurate predictions, while fewer outliers reflect greater consistency and overall reliability. In point cloud registration, these properties directly affect accuracy and usability, while in trajectory approximation, they signify how well NAR-based models mimic the algorithmic steps.

\textbf{Prediction Accuracy.}
We compare the predicted transformed point clouds from NAR-*ICP and the output of the algorithms against the ground truth, by calculating the \gls{mse}$(\downarrow)$ scores. To retrieve the transformed point clouds from the algorithm and the ground truth, we transform the source point cloud $P_m$ using the transformation matrices estimated by each. To then evaluate our models' performance in identifying correct registration correspondences, we employ several key classification metrics $(\uparrow)$: F1 score (F1), Precision (P), Recall (R), and Balanced Accuracy (A), similar to \cite{mdgat}. 



\begin{table*}[h!]
    \centering
    \caption{Registration performance comparison, in RTE$^{\textbf{GT}}$$(\downarrow)$ and RRE$^{\textbf{GT}}$$(\downarrow)$, of the final step of the baseline algorithms and the final step of the neural execution, against the ground truth from the dataset. Across almost all datasets and benchmarks, our NAR-based models outperform the algorithms they were trained on, demonstrating strong generalisation capabilities.}
    \renewcommand{\arraystretch}{1.4}
        \setlength{\tabcolsep}{10pt}

    \small
    \begin{tabular}{c|cc||cc|cc|cc}
         & \multicolumn{2}{c||}{Synthetic} & \multicolumn{2}{c|}{KITTI @ \SI{1.6}{\metre}} & \multicolumn{2}{c|}{KITTI @ \SI{11.3}{\metre}} & \multicolumn{2}{c}{KITTI @ \SI{24}{\metre}} \\
         Method & RTE & RRE & RTE & RRE & RTE & RRE & RTE & RRE \\
         \hline
         \texttt{P2P-ICP}   & 0.859 & 1.654 & \textbf{0.402} & \textbf{0.800} & 1.003 & 2.007 & 0.934 & 1.912 \\
         \texttt{NAR-P2Pv1}     & 0.904 & \textbf{1.585} & 0.432 & 0.843 & 0.828 & 1.746 & \textbf{0.888} & \textbf{1.769} \\
         \texttt{NAR-P2Pv2}     & \textbf{0.822} & 1.798 & 0.651 & 1.127 & \textbf{0.769} & \textbf{1.643} & 0.894 & \textbf{1.769} \\ \hline
         \texttt{P2L-ICP}       & 0.986 & 2.092 & 0.446 & 0.993 & 1.147 & 2.043 & 1.032 & 2.050 \\
         \texttt{NAR-P2L}       & \textbf{0.346} & \textbf{0.626} & \textbf{0.177} & \textbf{0.345} & \textbf{0.496} & \textbf{0.785} & \textbf{0.391} & \textbf{0.796} \\ \hline
         \texttt{G-ICP} & \textbf{0.901} & 2.063 & 0.518 & 1.232 & 1.022 & 2.168 & 0.944 & 1.981 \\
         \texttt{NAR-GICP} & 1.037 & \textbf{2.022} & \textbf{0.550} & \textbf{1.131} & \textbf{0.857} & \textbf{1.789} & \textbf{0.616} & \textbf{1.628} \\
    \end{tabular}
    \label{tab:y_good}
\end{table*}

\begin{table*}[h!]
    \centering
        \caption{Registration performance comparison, in RTE$^{\textbf{GT}}$$(\downarrow)$ and RRE$^{\textbf{GT}}$$(\downarrow)$, of the final step of the baseline algorithms and the final step of the neural execution after ground truth optimisation against the ground truth from the dataset. Our optimisation step significantly improves the registration performance of the underlying algorithm.}
    \small
    \renewcommand{\arraystretch}{1.4}
        \setlength{\tabcolsep}{10pt}
    \begin{tabular}{c|cc||cc|cc|cc}
         & \multicolumn{2}{c||}{Synthetic} & \multicolumn{2}{c|}{KITTI @ \SI{1.6}{\metre}} & \multicolumn{2}{c|}{KITTI @ \SI{11.3}{\metre}} & \multicolumn{2}{c}{KITTI @ \SI{24}{\metre}} \\
         Method & RTE & RRE & RTE & RRE & RTE & RRE & RTE & RRE \\
         \hline
        \texttt{P2P-ICP}   & 0.859 & 1.654 & 0.402 & 0.800 & 1.003 & 2.007 & 0.934 & 1.912 \\
         \texttt{NAR-P2Pv1$^+$} & \textbf{0.246} & \textbf{0.496} & \textbf{0.138} & \textbf{0.245} & 0.287 & 0.574 & 0.261 & 0.496 \\
         \texttt{NAR-P2Pv2$^+$} & 0.267 & 0.531 & 0.156 & 0.302 & \textbf{0.204} & \textbf{0.386} & \textbf{0.241} & \textbf{0.480} \\ \hline
        \texttt{P2L-ICP}       & 0.986 & 2.092 & 0.446 & 0.993 & 1.147 & 2.043 & 1.032 & 2.050 \\
         \texttt{NAR-P2L$^+$}   & \textbf{0.274} & \textbf{0.533} & \textbf{0.127} & \textbf{0.221} & \textbf{0.279} & \textbf{0.542} & \textbf{0.274} & \textbf{0.493} \\ \hline
        \texttt{G-ICP} & 0.901 & 2.063 & 0.518 & 1.232 & 1.022 & 2.168 & 0.944 & 1.981 \\
         \texttt{NAR-GICP$^+$}  & \textbf{0.292} & \textbf{0.579} & \textbf{0.175} & \textbf{0.328} & \textbf{0.276} & \textbf{0.556} & \textbf{0.222} & \textbf{0.458} \\
    \end{tabular}
    \label{tab:y_gt_rte_rre}
\end{table*}

\subsection{Training Details}

To finalise the training configuration, we compared various network architectures, \textit{teacher-forcing} probabilities, and hint configurations, as outlined in \cref{sec:ablation}. We used \textit{teacher-forcing} with a probability of $P_T=0.1$ to stabilise the trajectories, softmax for the categorical hints, and sigmoids for the mask hints. For the synthetic dataset, we generated $1000$/$64$/$64$ samples for training/evaluation/testing, respectively. As the encoder requires fixed-length inputs, we adjust input nodes by either randomly repeating them (padding) or randomly masking them. When comparing against the algorithms, the real-world dataset was split using a \num{60}/\num{20}/\SI{20}{\%} split across SemanticKITTI sequences $00$, $02$, $04$, and $06$, chosen for the route and environment variability in each. When comparing against the learned registration baselines, our evaluation follows the experimental setup proposed by GeoTransformer \cite{geotransformer}, including their data splits and scan distance settings. We train on SemanticKitti \cite{semantickitti} sequences $00-05$, evaluate on sequences $06-07$, and test on sequences $08-07$. 

Models were trained for \num{10000} steps with a batch size of $8$, calculating the loss for each intermediate hint and output. The hidden size was set to $256$, and the Adam optimiser was used with a learning rate of \num{1e-3}. Our models and learned baselines were trained on an NVIDIA RTX A6000 GPU, with all algorithmic computations and dataset generation running on the same GPU. Experiments were implemented using JAX, Haiku \cite{haiku2020github}, and the CLRS library \cite{clrs}.

\begin{table*}[h!]
    \centering
    \caption{Registration performance comparison, in RTE$^{\textbf{T}}$$(\downarrow)$ and RRE$^{\textbf{T}}$$(\downarrow)$, of our methods when learning to approximate the final step of the algorithm. \texttt{NAR-P2Pv1} and \texttt{NAR-P2Pv2} mostly outperform the more complex methods.}
    \small
    \renewcommand{\arraystretch}{1.4}
    \setlength{\tabcolsep}{10pt}
    \begin{tabular}{c|cc||cc|cc|cc}
         & \multicolumn{2}{c||}{Synthetic} & \multicolumn{2}{c|}{KITTI @ \SI{1.6}{\metre}} & \multicolumn{2}{c|}{KITTI @ \SI{11.3}{\metre}} & \multicolumn{2}{c}{KITTI @ \SI{24}{\metre}} \\
         Method & RTE & RRE & RTE & RRE & RTE & RRE & RTE & RRE \\
         \hline
         \texttt{NAR-P2Pv1} & \textbf{0.607} & \textbf{1.142} & \textbf{0.393} & \textbf{0.588} & \textbf{0.660} & \textbf{1.135} & \textbf{0.545} & \textbf{1.107} \\
         \texttt{NAR-P2Pv2} & \underline{0.637} & \underline{1.268} & 0.514 & \underline{0.781} & \underline{0.823} & \underline{1.447} & \underline{0.713} & \underline{1.383} \\
         \texttt{NAR-P2L}   & 0.952 & 2.066 & \underline{0.464} & 1.029 & 1.254 & 2.057 & 1.114 & 2.102 \\
         \texttt{NAR-GICP}  & 0.890 & 1.641 & 0.485 & 0.876 & 0.947 & 1.681 & 0.875 & 1.578  \\
    \end{tabular}
    \label{tab:y_icp_rte_rre}  
\end{table*}


\section{Results}
\label{sec:results}

We compare the performance of our models with baseline algorithms and learned  registration benchmarks. From \texttt{P2P-ICP} we learn two policies, \texttt{NAR-P2Pv1} and \texttt{NAR-P2Pv2}, from \texttt{P2L-ICP} we learn \texttt{NAR-P2L}, and from \texttt{G-ICP}, \texttt{NAR-GICP}. 
These correspond to the neural execution of each algorithm, respectively. When the method is superscripted with a $+$, such as \texttt{NAR-GICP$^+$}, the model was trained on the additional ground-truth optimisation step. For the neural execution of \texttt{P2P-ICP}, \texttt{NAR-P2Pv1} and \texttt{NAR-P2Pv2} differ in training strategy. \texttt{NAR-P2Pv2} is trained to approximate the two phases of the algorithm, separately. In comparison, \texttt{NAR-P2Pv1} uses a simpler, single-phase training strategy, approximating only the output from the second phase of the algorithm at each iteration. While it does not explicitly capture the internal computations of \texttt{P2P-ICP}, it offers a simpler architectural alternative that is easier to integrate within larger learning systems. We include \texttt{NAR-P2Pv1} as an interesting baseline to evaluate its performance against \texttt{NAR-P2Pv2}, and to validate both architectural alternatives. However, our primary interest lies in models like \texttt{NAR-P2Pv2}, which aim to approximate the full algorithmic trajectory and internal reasoning process. As such, the other methods, \texttt{NAR-P2L} and \texttt{NAR-GICP}, follow the two-phase approach, similar to \texttt{NAR-P2Pv2}, ensuring a more comprehensive representation of the algorithm's trajectory.

In the following, we compare the approaches both quantitatively and qualitatively against each other and the baselines. We evaluate our method in solving the registration task and in approximating the underlying algorithms.

\subsection{Registration: Quantitative Results}
\textbf{Comparison with Baseline Algorithms. }
To evaluate the performance of our point cloud registration method, we assess the \gls{rte} and \gls{rre} of different combinations of outputs and target signals, as defined in \cref{sec:metrics}. In \cref{tab:y_good}, we compare the baseline algorithms and their NAR-based execution against the ground truth, in RTE$^{\textbf{GT}}$ and RRE$^{\textbf{GT}}$. In particular, to demonstrate robustness under increasing relative transformations, we evaluate registration performance in the KITTI dataset across point cloud pairs separated by $\SI{1.6}{\metre}$, $\SI{11.3}{\metre}$, and $\SI{24}{\metre}$ on average. We observe that the performance of the algorithms degrades significantly from $\SI{1.6}{\metre}$ to $\SI{11.3}{\metre}$ and $\SI{24}{\metre}$, while, even though our NAR-*ICP methods show some decline, they remain consistently more accurate. 

Our results indicate that the NAR-based execution outperforms the algorithm it was trained to approximate when comparing both with the ground truth. This is mainly because, unlike traditional \gls{icp}-based algorithms, which are highly sensitive to a good initial alignment, our model learns the initialisation, making it more robust to large initial transformations. Additionally, the NAR-*ICP framework learns a parametrised reasoning process that generalises well across data distributions, handling noise, partial observations, and complex geometries. The model also does not assume a perfect expert; instead, it leverages the algorithm's structure as supervision while learning to improve upon it through convergence tracking and learned termination. As a result, NAR-*ICP performs more robustly and reliably across a wide range of challenging registration scenarios.

We additionally compare the performance of the neural execution after ground truth optimisation, as shown in \cref{tab:y_gt_rte_rre}, to demonstrate the performance enhancement of our method. Here, even though we slightly diverge from the original algorithm, we retain its core computational properties while improving its accuracy. Notably, the registration performance across all benchmarks improves significantly across the board in our testing datasets, as reflected in \cref{tab:y_gt_rte_rre}, compared to training without this step in \cref{tab:y_good}. This improvement is particularly significant for the point cloud registration task, as the optimisation step not only approximates the original algorithm but also substantially boosts its performance.

Lastly, our results in \cref{tab:y_icp_rte_rre} indicate that the performance of \texttt{NAR-P2Pv1} and \texttt{NAR-P2Pv2} is comparable when approximating the output of the algorithms, in RTE$^{\textbf{T}}$ and RRE$^{\textbf{T}}$, while both outperform the other benchmarks. The comparison between \texttt{NAR-P2Pv1} and \texttt{NAR-P2Pv2} in \cref{tab:y_good}, \cref{tab:y_gt_rte_rre}, and \cref{tab:y_icp_rte_rre} reveals an interesting trade-off between performance and introspection. While \texttt{NAR-P2Pv1} achieves slightly better registration accuracy in some of our experiments, \texttt{NAR-P2Pv2} approximates a more complete algorithmic trajectory, effectively emulating the internal computations of \texttt{ICP-P2P}. As a result, \texttt{NAR-P2Pv2} serves as a more intricate and, therefore, interesting benchmark compared to \texttt{NAR-P2Pv1}. Despite not approximating the internal algorithmic computations, \texttt{NAR-P2Pv1} remains a valuable baseline as it is learning to approximate the direct algorithmic outputs at each iteration.

\begin{figure*}[ht!]
    \centering
    \captionsetup{justification=centering}
    \begin{subfigure}{0.95\textwidth}
    \centering
        \includegraphics[trim=280 40 230 97,clip,width=0.15\textwidth]{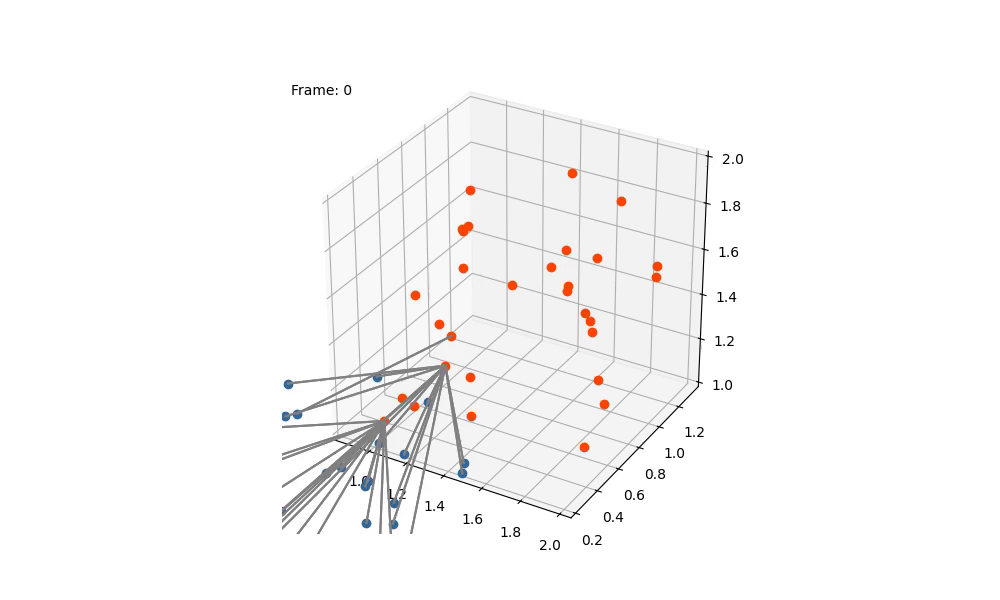}
        \includegraphics[trim=280 40 230 97,clip,width=0.15\textwidth]{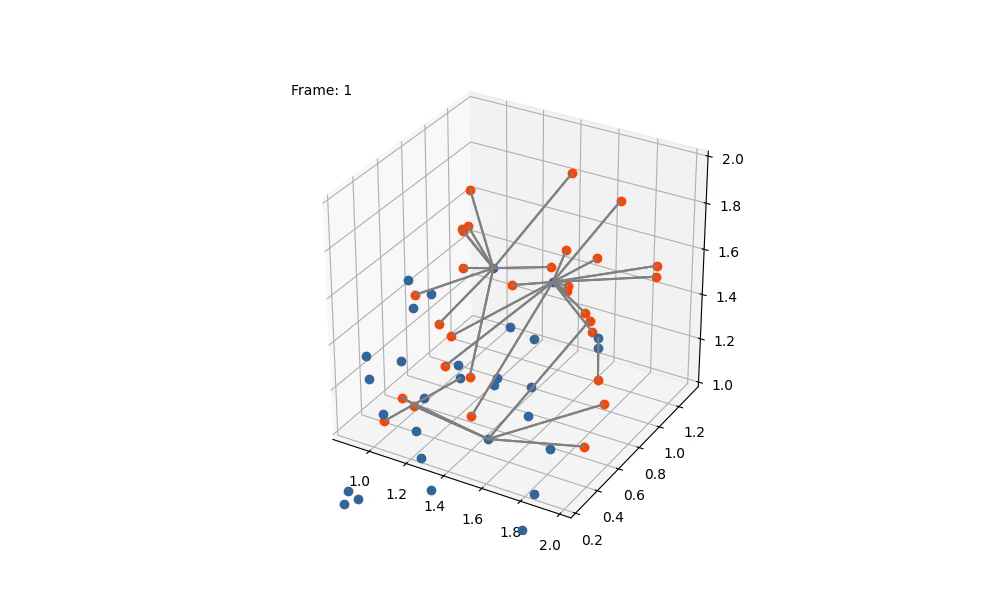}
        \includegraphics[trim=280 40 230 97,clip,width=0.15\textwidth]{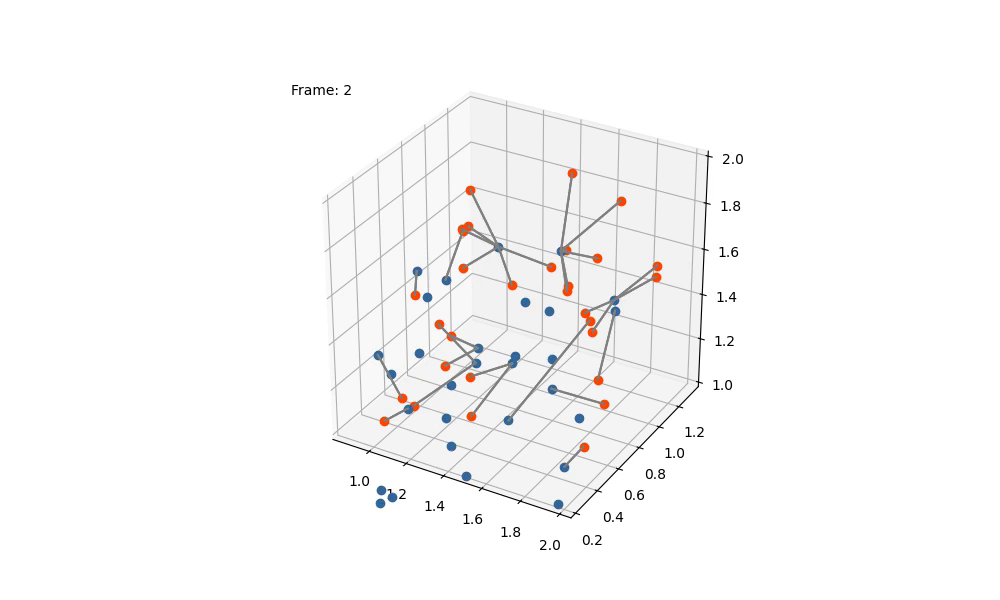}
        \includegraphics[trim=280 40 230 97,clip,width=0.15\textwidth]{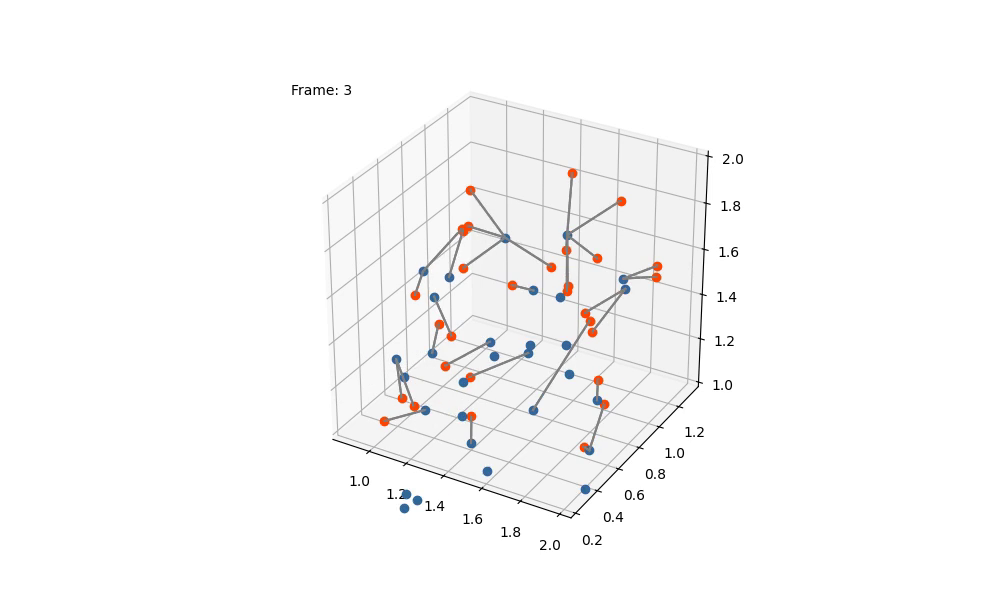}
        \includegraphics[trim=280 40 230 97,clip,width=0.15\textwidth]{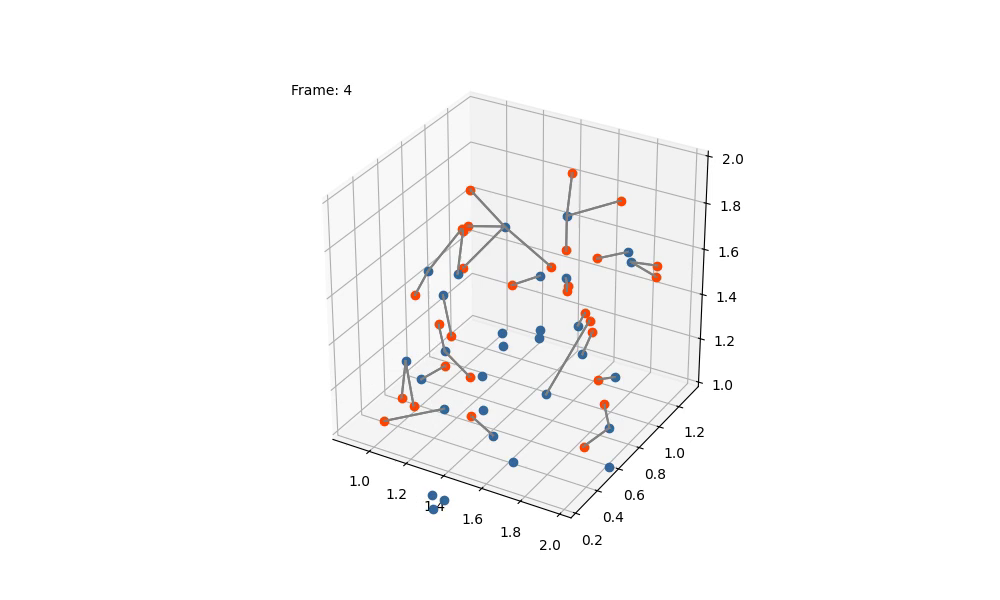}
        \includegraphics[trim=280 40 230 97,clip,width=0.15\textwidth]{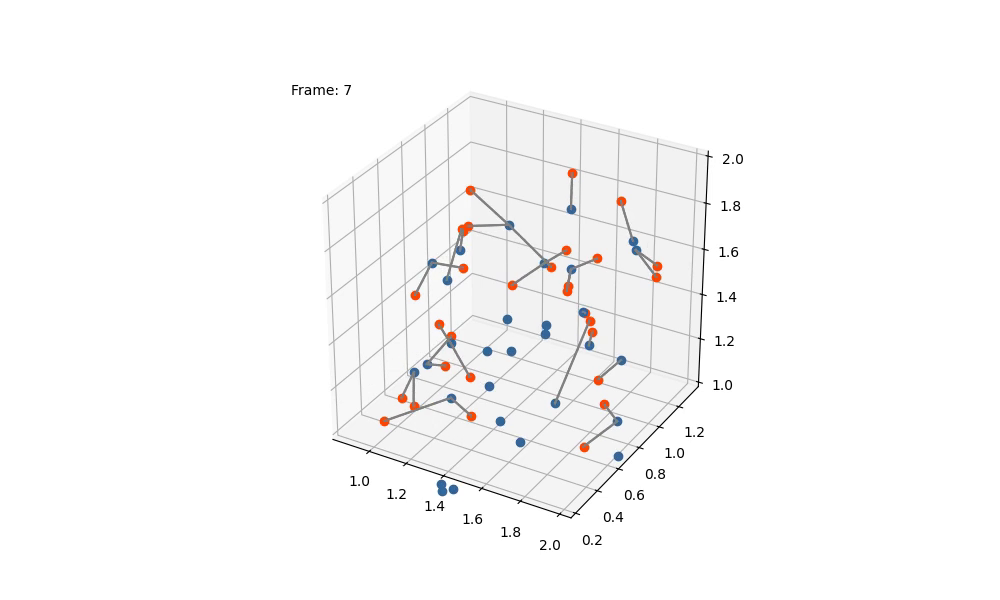}
        \caption{}
    \end{subfigure}
    
        \begin{subfigure}{0.95\textwidth}
        \centering
        \includegraphics[trim=280 40 230 97,clip,width=0.15\textwidth]{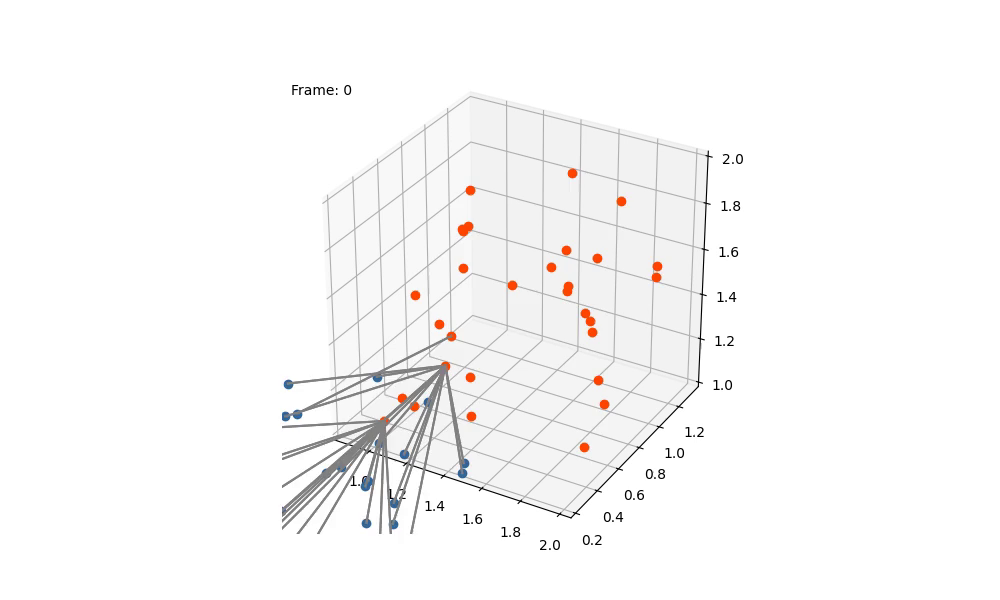}
        \includegraphics[trim=280 40 230 97,clip,width=0.15\textwidth]{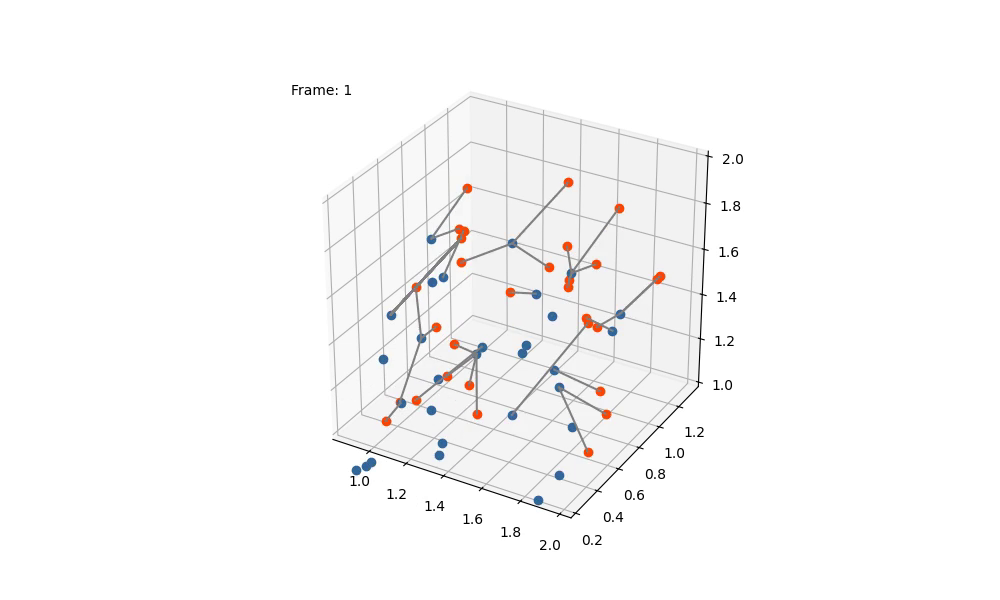}
        \includegraphics[trim=280 40 230 97,clip,width=0.15\textwidth]{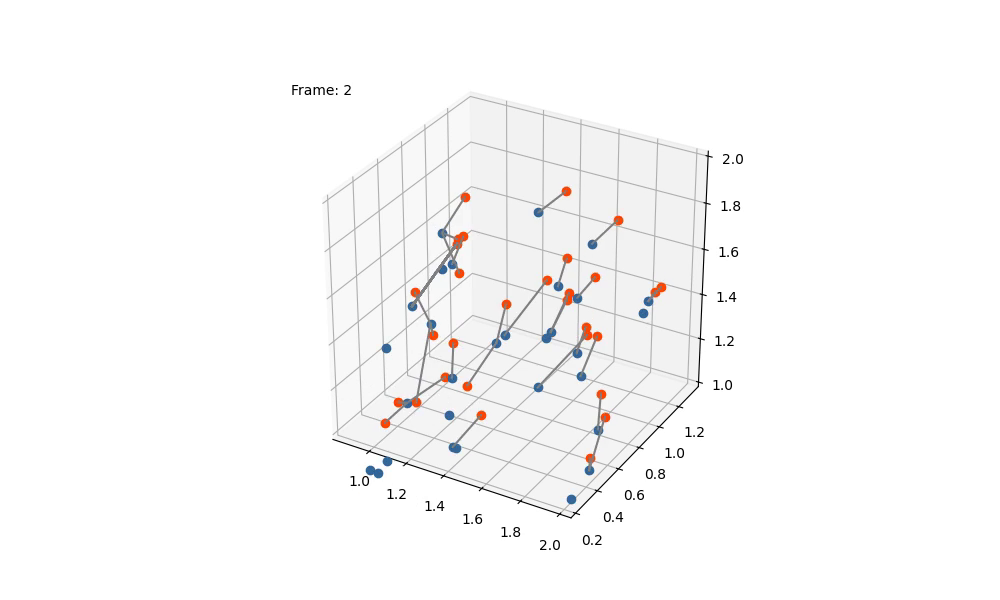}
        \includegraphics[trim=280 40 230 97,clip,width=0.15\textwidth]{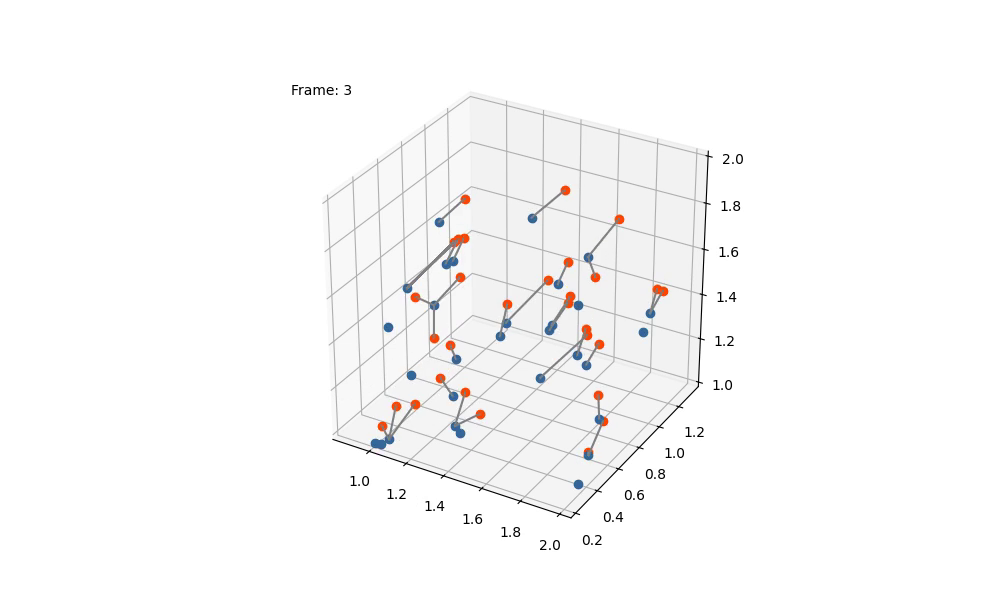}
        \includegraphics[trim=280 40 230 97,clip,width=0.15\textwidth]{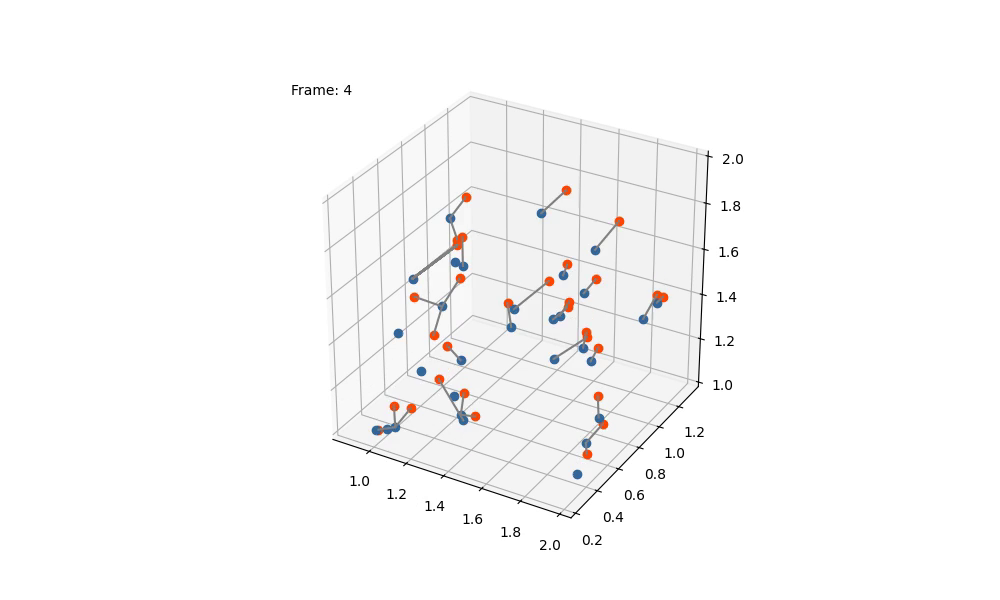}
        \includegraphics[trim=280 40 230 97,clip,width=0.15\textwidth]{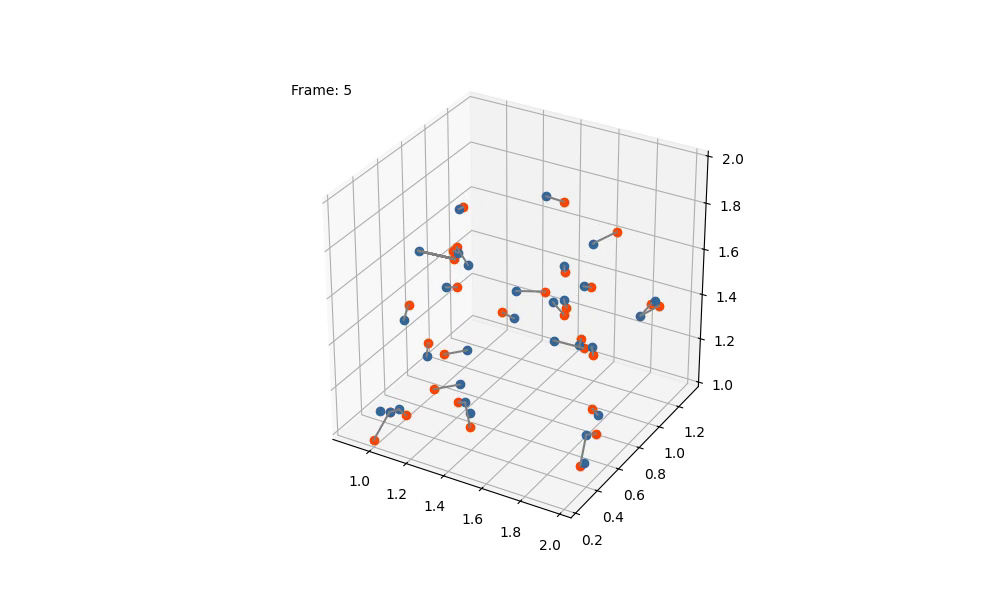}
        \caption{}
    \end{subfigure}
    \captionsetup{justification=justified}
    \caption{Visualisation of the intermediate steps output for (a) \texttt{P2P-ICP} and (b) ours, \texttt{NAR-P2Pv2}. Despite being trained on the latter, \texttt{NAR-P2Pv2} demonstrates superior registration performance, achieving better point cloud alignment and finding more accurate correspondences. To simplify the visualisation, we use the predictions for the $\textcolor{myblue}{\mathtt{phase}}$ hint to identify the intermediate algorithmic components and display those from the second phase of the algorithm.}
    \label{fig:intermediate}
    \vspace{-5pt}
\end{figure*}

\textbf{Comparison with Learned Baselines. }We compare our method against learned \gls{icp}-based and state-of-the-art registration baselines, described in \Cref{subsec:learnedbaselines}, using standard metrics: RTE and RRE, as before, and Registration Recall (RR, in \%). RR corresponds to the percentage of successfully aligned scans for RTE$<2m$ and RRE$<0.6^\circ$. 

Overall, all our NAR-*ICP models achieve superior or on-par performance compared to the baselines, across all metrics, as seen in \cref{tab:benchmarks}, with \texttt{NAR-GICP$^+$} standing out as the best-performing overall. Specifically, \texttt{NAR-P2L$^+$} achieves the best RTE, followed by \texttt{NAR-GICP$^+$} and \texttt{NAR-P2Pv2$^+$}, demonstrating substantial improvement over state-of-the-art methods like \texttt{GeoTransformer} and \texttt{Predator}. For RRE, \texttt{NAR-GICP$^+$} and \texttt{NAR-P2Pv2$^+$} significantly outperform the baselines, which is particularly notable since RRE is often the most challenging metric on SemanticKITTI. In RR, both \texttt{NAR-GICP$^+$} and \texttt{NAR-P2Pv2$^+$} methods achieve strong results and comparable performance to the best performers. Although \texttt{DCP} achieves the best recall, our models offer better balance, with consistently lower translation and rotation errors.

These results reinforce the core message of our work; while our primary contribution is to demonstrate how algorithmic reasoning can be integrated into neural networks for point cloud registration, our approach also achieves strong performance as a learned registration method.

\begin{table}[h]
\caption{Comparison with state-of-the-art learned registration baselines on the SemanticKITTI dataset, in RTE ($\downarrow$), RRE ($\downarrow$), RR($\uparrow$). Best results are marked in \textbf{bold}, second-best are \underline{underlined}, and the overall best is highlighted.}
\centering
\renewcommand{\arraystretch}{1.4}
\resizebox{0.9\columnwidth}{!}{
\begin{tabular}{l||l|ccc}
 & Method       & RTE$^{\textbf{GT}}$ & RRE$^{\textbf{GT}}$ & RR$^{\textbf{GT}}$  \\ \hline
\multirow{5}{*}{\rotatebox{90}{Baselines}} &\texttt{GeoTransformer}    & 0.335  & 0.512 & 85.2      \\
 &\texttt{Predator} & 0.433   & 0.371     & 74.1      \\
 &\texttt{DiffICP}     & 0.245    & 0.375     & \underline{98.7}    \\
 &\texttt{DGR}      & 0.252   & 0.371     & \underline{98.7}      \\
 &\texttt{DCP}      & 0.147   & 0.376   &  \textbf{99.4}         \\
 \hline
\multirow{4}{*}{\rotatebox{90}{\textit{Ours}}} & \texttt{NAR-P2Pv1$^+$} & 0.148 & 0.660  & 96.8   \\
 & \texttt{NAR-P2Pv2$^+$} & 0.148   & \underline{0.334}     & 98.2     \\
 & \texttt{NAR-P2L$^+$} & \textbf{0.135} & 0.391  & 97.1  \\
 & \cellcolor[HTML]{FFFFE8}\texttt{NAR-GICP$^+$} & \cellcolor[HTML]{FFFFE8}\underline{0.137} & \cellcolor[HTML]{FFFFE8}\textbf{0.333}  & \cellcolor[HTML]{FFFFE8}98.2  
\end{tabular}
}
\vspace{-5pt}
\label{tab:benchmarks}
\end{table}

\subsection{Registration: Qualitative Results}
We visualise the registration performance of the neural algorithmic execution in comparison to the baseline algorithm in \Cref{fig:intermediate}. Here, we use the same set of point clouds from the testing set of the synthetic dataset and demonstrate that \texttt{NAR-P2Pv2} achieves smoother registration performance than the baseline algorithm in the challenging synthetic dataset, highlighting its strong generalisation capabilities. Additionally, while \texttt{P2P-ICP} requires an accurate first initialisation step, making it sensitive to the initial guess, \texttt{NAR-P2Pv2} learns the initialisation as part of the algorithmic trajectory, leading to more reliable results without the need for this first manual estimation. By learning to infer optimal initialisations, NAR-*ICP removes a key limitation of traditional ICP-based algorithms. Notably, the visualisation illustrates that while the baseline \texttt{P2P-ICP} struggles to find strong correspondences and converge, \texttt{NAR-P2Pv2} -- despite being trained on \texttt{P2P-ICP} -- successfully identifies strong correspondences and achieves more accurate point cloud registration than the algorithm. These results highlight the effectiveness and promise of our approach in addressing challenging robotics tasks using neural networks while maintaining the compositionality and interpretability of the algorithms.

\subsection{Runtime Performance}\label{subsec:runtime}
\textbf{Learned Baselines. } \Cref{tab:learned_runtime_params_transposed} shows the runtime-parameter trade-offs across our NAR-*ICP models and our learned registration baselines. 
For iterative methods, runtime is measured over the full sequence of iterations until convergence, i.e., until the predefined stopping criterion is met. All NAR-*ICP variants demonstrate strong computational efficiency, with \texttt{NAR-P2Pv2$^+$} achieving the fastest inference time, while even the more complex \texttt{NAR-P2L$^+$} and \texttt{NAR-GICP$^+$} remain faster than almost all baselines. It is worth noting that our iterative models are significantly faster than all non-iterative baselines, which makes the comparison even more interesting. Additionally, nearly all benchmarks are considerably more complex in terms of parameter count. The only exception is \texttt{DiffICP}, which we deliberately implemented as a small model for comparison. Despite its fewer parameters, its runtime is slower than our NAR-*ICP models, suggesting that model architecture also plays a crucial role in inference efficiency. Our NAR-*ICP models stand out as the second most lightweight option, drastically reducing the number of parameters required, compared to baselines such as \texttt{DGR} and \texttt{Predator}. Among them, \texttt{NAR-P2Pv2$^+$} achieves the fastest inference, outperforming the most efficient baseline, \texttt{DGR}, while using significantly fewer parameters (approximately $7\times$ less).
These results demonstrate that our methods achieve a good balance between architecture and inference speed. Unlike large baselines, which rely on heavy models, our approach achieves efficient inference with a lightweight design, and unlike \texttt{DiffICP}, this efficiency comes from the architecture rather than parameter count alone. 

\begin{table*}[t]
\centering
\caption{Runtime comparison in terms of average inference time (in seconds) per pair of scans and number of network parameters of our NAR-*ICP models against learned baselines. Method names have been shortened for brevity.}
\small
\renewcommand{\arraystretch}{1.4}
\begin{tabular}{l|ccccc||cccc}
 & \parbox{1cm}{\centering\texttt{GeoTr}} & \parbox{1cm}{\centering\texttt{Pred}} & \parbox{1cm}{\centering\texttt{Diff}} & \parbox{1cm}{\centering\texttt{DGR}} & \parbox{1cm}{\centering\texttt{DCP}} & \parbox{1.4cm}{\centering\texttt{N-P2Pv1}} & \parbox{1.4cm}{\centering\texttt{N-P2Pv2}} & \texttt{N-P2L} & \texttt{N-GICP} \\ \hline
Time (s) ($\downarrow$) &
0.13 & 0.34 & 0.10 & 0.09 & \underline{0.03} &
\underline{0.03} & \textbf{0.02} & 0.07 & 0.08 \\ \hline
Parameters ($\downarrow$) &
1M & 22.8M & \textbf{25k} & 235M & 5.56M &
\multicolumn{4}{c}{\texttt{NAR-$^*$ICP:} \underline{773k}} \\
\end{tabular}
\label{tab:learned_runtime_params_transposed}
\vspace{-5pt}
\end{table*}

\textbf{Baseline Algorithms. } We compare the total runtime and the average number of steps for each algorithm and its corresponding neural execution, with (stop) and without (last) termination, as shown in \cref{tab:time}. Here we note that the version of our models without learned termination - i.e. NAR-*ICP (last) - is equivalent to the \texttt{NAR-CLRS} baseline, as described in \Cref{subsec:clrs30nar}. Through these experiments, we assess runtime and step efficiency not only against the algorithms but also against the \texttt{NAR-CLRS} baseline, where termination is not learned. The average number of steps is determined using the $\textcolor{myblue}{\mathtt{phase}}$ hint, which predicts when the algorithm enters its second phase. Notably, the NAR-*ICP models reduce the runtime of \texttt{P2L-ICP} and \texttt{G-ICP}, which, as they require additional computations, such as the estimation of normals and covariances, naturally result in longer execution times. Our models address and mitigate this drawback, leading to improved efficiency. 
Additionally, in almost all cases, our method requires fewer execution steps to achieve the desired output compared to the baseline algorithms. Overall, our learned termination (stop) significantly improves average runtime and execution steps (more than $2\times$ reduction) compared to \texttt{NAR-CLRS}, as seen in the NAR-*ICP (last) models. These position our models as good alternatives compared to traditional \gls{icp}-based algorithms, especially in time-sensitive applications. 

\setlength{\tabcolsep}{2.3pt}
\begin{table}[]
\centering
\caption{Runtime performance and average number of steps comparison for each algorithm and their respective neural execution with (stop) and without (last) termination. Our termination network significantly improves the total runtime and the average number of steps required for our models.}
\small
\renewcommand{\arraystretch}{1.3}
\setlength{\tabcolsep}{7pt}
\begin{tabular}{c|c||c}
\multirow{2}{*}{} &  Time (s) ($\downarrow$) & \quad Avg. Steps ($\downarrow$) \\ \hline
\texttt{P2P-ICP}         & \textbf{0.051}   & \quad 10        \\ 
\texttt{NAR-P2Pv1$^+$} (last)        & 0.142 & \quad    15     \\
\texttt{NAR-P2Pv1$^+$} (stop)        & 0.064 & \quad \textbf{7}        \\ \hline
\texttt{P2P-ICP}         & \textbf{0.051}   & \quad \textbf{10}        \\ 
\texttt{NAR-P2Pv2$^+$} (last)        & 0.172 & \quad  37   \\ 
\texttt{NAR-P2Pv2$^+$} (stop)        &  0.069 & \quad 15       \\ 
\hline
\texttt{P2L-ICP}           & 0.078 & \quad 28        \\
\texttt{NAR-P2L$^+$} (last)        & 0.270 & \quad  98   \\ 
\texttt{NAR-P2L$^+$} (stop)          & \textbf{0.046} &  \quad \textbf{17}        \\ \hline
\texttt{G-ICP}         & 0.946 & \quad 32  \\ 
\texttt{NAR-GICP$^+$} (last)        & 0.268  & \quad   98  \\ 
\texttt{NAR-GICP$^+$} (stop)        &  \textbf{0.056} &  \quad \textbf{21}        \\ 
\end{tabular}
\label{tab:time}
\vspace{-5pt}
\end{table}

\begin{figure*}[t!]
    \centering
\includegraphics[width=0.95\textwidth]{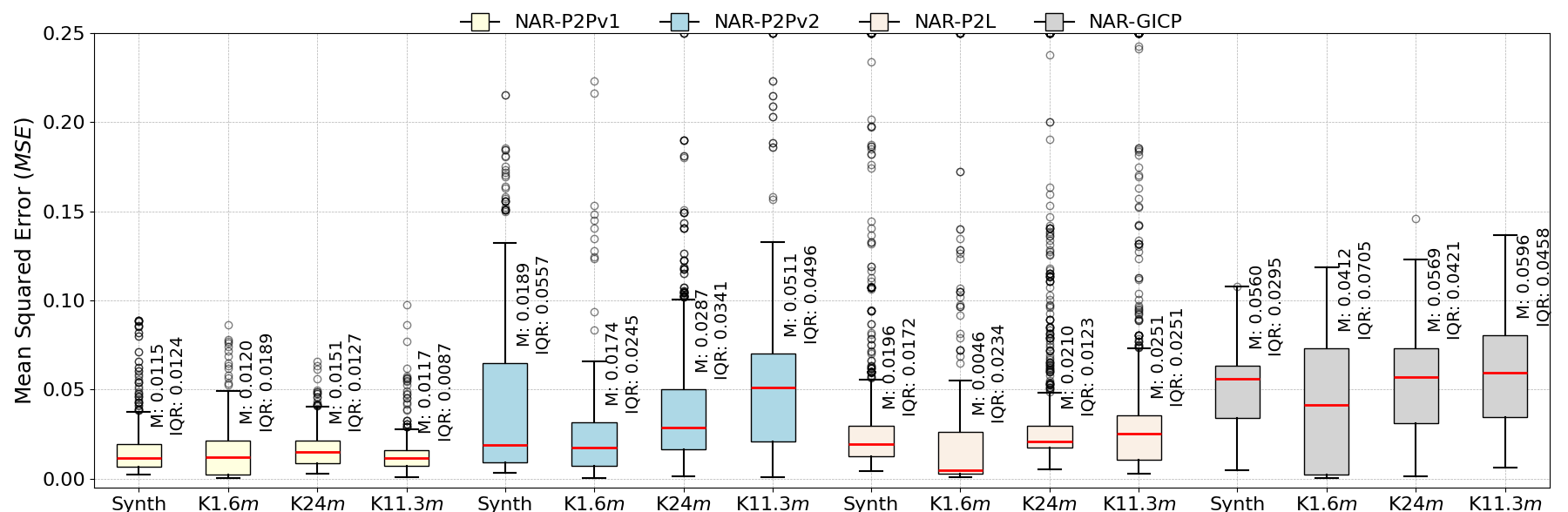}
\caption{Error distributions of the predicted transformed point clouds during the neural execution at each intermediate step across all benchmarks, in MSE$^\textbf{t}$$(\downarrow)$ (outliers capped at $0.25$ for clarity), median error, and IQR.}
\label{fig:intermediate_box_plots}
\vspace{5pt}
\end{figure*}

\subsection{Algorithmic Alignment } To assess the performance of our method in approximating the algorithms, we extract the MSE$^\textbf{t}$ error for the intermediate algorithmic steps, as shown in \Cref{fig:intermediate_box_plots}. As expected, \texttt{NAR-P2Pv1} consistently outperforms the other benchmarks as it learns a simpler algorithmic trajectory. Among the more complex benchmarks (\texttt{NAR-P2Pv2}, \texttt{NAR-P2L}, and \texttt{NAR-GICP}), \texttt{NAR-P2L} achieves the lowest medians and IQRs, indicating better model performance in terms of error variance. However, the significant number of outliers in \texttt{NAR-P2L} suggests that the model is far less reliable due to its extreme prediction errors. In contrast, \texttt{NAR-GICP} has almost zero outliers, resulting in consistent and stable predictions, which is preferable for our task. \texttt{NAR-P2Pv2} demonstrates low medians and IQRs, suggesting good overall performance and reduced error variability, but the presence of outliers in some datasets makes it less reliable than \texttt{NAR-GICP}. Notably, \texttt{NAR-GICP} stands out as the most reliable benchmark for challenging datasets with larger relative displacements and lower point cloud overlap, whereas \texttt{NAR-P2Pv2} excels in scenarios with smaller relative displacements. 

\setlength{\tabcolsep}{2.5pt}
\begin{table*}[]
\centering
\caption{\gls{mse}$^{\textbf{T}}(\times10^{-2})$ $(\downarrow)$ of the predicted final transformed point clouds. \texttt{NAR-P2Pv1}, \texttt{NAR-P2Pv2}, and \texttt{NAR-P2L} achieve comparable predictive performance across the datasets.}
\small
\renewcommand{\arraystretch}{1.4}
\setlength{\tabcolsep}{3pt}
\begin{tabular}{c|ccc||ccc|ccc|ccc}
\multicolumn{1}{l|}{\multirow{2}{*}{}}   & \multicolumn{3}{c||}{Synthetic} & \multicolumn{3}{c|}{KITTI @ \SI{1.6}{\metre}}  & \multicolumn{3}{c|}{KITTI @ \SI{11.3}{\metre}}   & \multicolumn{3}{c}{KITTI @ \SI{24}{\metre}}  \\ 
 Method    & $MSE_x$   & $MSE_y$   & $MSE_z$  & $MSE_x$ & $MSE_y$ & $MSE_z$  & $MSE_x$  & $MSE_y$    & $MSE_z$  & $MSE_x$  & $MSE_y$ & $MSE_z$ \\ \hline
\texttt{NAR-P2Pv1}   & \textbf{1.792} & 2.386 & 2.267 & 1.114 & 0.744 & \textbf{0.619} & 1.970 & 1.824 & \textbf{1.718} & 2.986 & 2.138 & \textbf{1.682} \\
\texttt{NAR-P2Pv2}   & 5.026 & 4.820 & 4.326 & \textbf{0.562} & \textbf{0.370} & 1.131 & \textbf{1.777} & \textbf{1.604} & 2.334 & 4.828 & \textbf{1.508} & 1.935 \\
\texttt{NAR-P2L}   & 1.903 & \textbf{1.968} & \textbf{1.706} & 1.055 & 0.975 & 0.799 & 3.052 & 2.367 & 2.727 & \textbf{2.132} & 2.970 & 1.999 \\
\texttt{NAR-GICP}   & 4.924 & 5.088 & 5.111 & 3.139 & 2.546 & 2.830 & 5.346 & 4.547 & 5.489 & 4.083 & 5.202 & 4.821
\end{tabular}
\label{tab:mse_pred}
\end{table*}

\begin{table*}[]
\centering
\caption{Classification scores $(\uparrow)$, F1$^{\textbf{T}}$, P$^{\textbf{T}}$, R$^{\textbf{T}}$, and A$^{\textbf{T}}$, of the predicted final correspondences. \texttt{NAR-P2Pv1} demonstrates superior performance on the synthetic dataset, while \texttt{NAR-P2Pv2} on the real-world datasets.}
\small
\renewcommand{\arraystretch}{1.4}
\setlength{\tabcolsep}{5pt}
\begin{tabular}{c|cccc||cccc|cccc|cccc}
 & \multicolumn{4}{c||}{Synthetic}   & \multicolumn{4}{c|}{KITTI @ \SI{1.6}{\metre}} & \multicolumn{4}{c|}{KITTI @ \SI{11.3}{\metre}}  & \multicolumn{4}{c}{KITTI @ \SI{24}{\metre}}   \\ 
\multicolumn{1}{c|}{Method} & \multicolumn{1}{c}{F1} & \multicolumn{1}{c}{A} & \multicolumn{1}{c}{P} & \multicolumn{1}{c||}{R} & \multicolumn{1}{c}{F1} & \multicolumn{1}{c}{A} & \multicolumn{1}{c}{P} & \multicolumn{1}{c|}{R} & \multicolumn{1}{c}{F1} & \multicolumn{1}{c}{A} & \multicolumn{1}{c}{P} & \multicolumn{1}{c|}{R} & \multicolumn{1}{c}{F1} & \multicolumn{1}{c}{A} & \multicolumn{1}{c}{P} & \multicolumn{1}{c}{R} \\ \hline
\texttt{NAR-P2Pv1} & \textbf{0.30} & \textbf{0.63} & \textbf{0.29} & \textbf{0.31} & 0.63 & 0.80 & 0.63 & 0.62 & 0.31 & 0.63 & 0.30 & 0.31 & 0.29 & 0.62 & 0.28 & 0.30 \\
\texttt{NAR-P2Pv2} & 0.18 & 0.56 & 0.17 & 0.19 & \textbf{0.78} & \textbf{0.88} & \textbf{0.79} & \textbf{0.77} & \textbf{0.39} & \textbf{0.69} & \textbf{0.38} & \textbf{0.42} & \textbf{0.42} & \textbf{0.69} & \textbf{0.43} & \textbf{0.40} \\
\texttt{NAR-P2L} & 0.09 & 0.52 & 0.10 & 0.09 & 0.52 & 0.75 & 0.53 & 0.52 & 0.10 & 0.52 & 0.10 & 0.10 & 0.11 & 0.52 & 0.11 & 0.10 \\
\texttt{NAR-GICP} & 0.16 & 0.56 & 0.15 & 0.17 & 0.47 & 0.72 & 0.48 & 0.47 & 0.11 & 0.53 & 0.11 & 0.11 & 0.20 & 0.57 & 0.20 & 0.20 \\
\end{tabular}
\label{tab:class_pred}
\vspace{-5pt}
\end{table*}

\subsection{Prediction Accuracy }
To further evaluate the performance of our methods in predicting algorithmic outputs, we compare the MSE$^{\textbf{T}}$ scores and classification metrics, F1$^{\textbf{T}}$, P$^{\textbf{T}}$, R$^{\textbf{T}}$, and A$^{\textbf{T}}$, across our datasets. Our results in \cref{tab:mse_pred} demonstrate that \texttt{NAR-P2Pv1}, \texttt{NAR-P2Pv2}, and \texttt{NAR-P2L} achieve comparable performance to each other in predicting the transformed point clouds. However, as shown in \cref{tab:class_pred}, \texttt{NAR-P2Pv2} outperforms the benchmarks in predicting the correspondences on the real-world datasets, while \texttt{NAR-P2Pv1} excels on the synthetic dataset. Notably, our methods achieve strong performance in the less challenging dataset, KITTI @ \SI{1.6}{\metre}, where the displacement between point clouds is smaller. 

\subsection{Optimisation of Neural Execution}
\textbf{Termination. }
We evaluate the effectiveness of our added termination criterion, $\textcolor{myblue}{\mathtt{stop}}$, in NAR-*ICP, in comparison to \texttt{NAR-CLRS} as described in \Cref{subsec:clrs30nar}. In CLRS-30, the number of hint probes directly dictates the number of processor iterations. Our evaluation, therefore, focuses on whether introducing a termination network improves prediction accuracy. We first assess the average \gls{mse} between the transformed point clouds at the termination step (stop) and the output of the algorithm, alongside the average \gls{mse} between the last step of \texttt{NAR-CLRS} and the output of the algorithm, as seen in \cref{tab:terminations}. As in \Cref{subsec:runtime}, NAR-*ICP (last) in our tables is equivalent to \texttt{NAR-CLRS}; we adopt this notation to facilitate comparisons. Our results demonstrate that our models with termination (stop) achieve superior or on-par performance across all datasets and benchmarks, compared to NAR-*ICP (last), i.e. \texttt{NAR-CLRS}. We note that runtime performance and average processor steps (with and without termination) have been compared in \ref{subsec:runtime} and \Cref{tab:time}, with results indicating notable performance improvements. Overall, by adding the termination, we not only reduce runtime but also significantly enhance the accuracy of the method.

\begin{table*}[]
\centering
\caption{Performance comparison at the predicted termination iteration and at the last iteration from \texttt{NAR-CLRS}, measured in average \gls{mse}$^{\textbf{T}}$. The output from our termination step consistently outperforms the output from the last step.}
\renewcommand{\arraystretch}{1.4}
\small
\setlength{\tabcolsep}{8pt}
\begin{tabular}{c|cc||cc|cc|cc}
 & \multicolumn{2}{c||}{Synthetic} & \multicolumn{2}{c|}{KITTI @ \SI{1.6}{\metre}}  & \multicolumn{2}{c|}{KITTI @ \SI{11.3}{\metre}}  & \multicolumn{2}{c}{KITTI @ \SI{24}{\metre}}  \\ 
Method & last & stop & last & stop & last & stop & last & stop \\ \hline
\texttt{NAR-P2Pv1}  & 0.028  & \textbf{0.021} & 0.018    & \textbf{0.008 }  & 0.033    & \textbf{0.018}   & 0.031   & \textbf{0.023}  \\
\texttt{NAR-P2Pv2}   & \textbf{0.043} & 0.047    & 0.018  & \textbf{0.007 }   & 0.053  & \textbf{0.019}    & 0.039 & \textbf{0.028}   \\
\texttt{NAR-P2L} &  \textbf{0.019}  & \textbf{0.019}  & \textbf{0.009 } & \textbf{0.009 } & \textbf{0.025} & 0.027   &\textbf{0.021} & 0.024   \\
\texttt{NAR-GICP}  & 0.054 & \textbf{0.050} & 0.029 & \textbf{0.028} & 0.056 & \textbf{0.051}  & 0.051 & \textbf{0.047} \\ 
\end{tabular}
\label{tab:terminations}
\end{table*}

\setlength{\tabcolsep}{2.5pt}
\begin{table*}[]
\centering
\caption{\gls{mse}$^{\textbf{GT}}(\times10^{-2})$ $(\downarrow)$ of the predicted final transformed point clouds after ground truth optimisation compared to the algorithmic output, against the ground truth. This step significantly improves the performance of the neural execution, surpassing the baseline algorithms across all datasets.}
\small
\renewcommand{\arraystretch}{1.4}
\setlength{\tabcolsep}{3pt}
\begin{tabular}{c|ccc||ccc|ccc|ccc}
\multicolumn{1}{l|}{\multirow{2}{*}{}}   & \multicolumn{3}{c||}{Synthetic} & \multicolumn{3}{c|}{KITTI @ \SI{1.6}{\metre}}  & \multicolumn{3}{c|}{KITTI @ \SI{11.3}{\metre}}   & \multicolumn{3}{c}{KITTI @ \SI{24}{\metre}}  \\ 
 Method    & $MSE_x$   & $MSE_y$   & $MSE_z$  & $MSE_x$ & $MSE_y$ & $MSE_z$  & $MSE_x$  & $MSE_y$    & $MSE_z$  & $MSE_x$  & $MSE_y$ & $MSE_z$ \\ \hline
\texttt{P2P-ICP}      & 6.568  & 7.874  & 4.469 &    3.268 & 2.524 & 2.743 & 9.288 & 10.516 & 4.361 & 8.822  & 8.697 & 4.225  \\
\texttt{NAR-P2Pv1$^+$}   & \textbf{1.101} & \textbf{1.169} & \textbf{0.072} &  \textbf{0.764} & \textbf{0.592} & \textbf{0.068} & \textbf{1.203} & \textbf{1.420} & \textbf{0.035} & \textbf{1.163} & \textbf{0.991} & \textbf{0.027}   \\
\hline
\texttt{P2P-ICP}      & 6.568  & 7.874  & 4.469 &    3.268 & 2.524 & 2.743 & 9.288 & 10.516 & 4.361 & 8.822  & 8.697 & 4.225  \\

\texttt{NAR-P2Pv2$^+$}  & \textbf{1.117} & \textbf{1.115} & \textbf{0.101} & \textbf{0.625} & \textbf{0.709} & \textbf{0.066} & \textbf{0.902} & \textbf{1.203} & \textbf{0.104} & \textbf{1.075} & \textbf{1.212} & \textbf{0.178}  \\
\hline
\texttt{P2L-ICP} & 4.655          & 4.924          & 4.724  &  2.281          & 1.888          & 2.129          & 4.444          & 4.598          & 5.031          & 4.484          & 5.461          & 4.783            \\

\texttt{NAR-P2L$^+$} & \textbf{1.355} & \textbf{1.354} & \textbf{0.088} &   \textbf{0.806} & \textbf{0.533} & \textbf{0.140} & \textbf{1.410} & \textbf{1.391} & \textbf{0.168} & \textbf{1.508} & \textbf{1.422} & \textbf{0.115}  
 \\
\hline
\texttt{G-ICP} & 10.106 & 9.911 & 5.727 & 6.498 & 4.889 & 3.980 & 9.327 & 11.122 & 7.104 & 9.250 & 9.712 & 5.964 \\

\texttt{NAR-GICP$^+$}   &  \textbf{1.302} & \textbf{1.157} & \textbf{0.065} & \textbf{0.662} & \textbf{0.655} & \textbf{0.037} & \textbf{1.587} & \textbf{1.412} & \textbf{0.275} & \textbf{0.913} & \textbf{0.848} & \textbf{0.139} \\
\end{tabular}
\label{tab:mse_optimisation}

\end{table*}

\begin{table*}[]
\centering
\caption{Classification scores $(\uparrow)$, F1$^{\textbf{GT}}$, P$^{\textbf{GT}}$, R$^{\textbf{GT}}$, and A$^{\textbf{GT}}$, of the predicted final correspondences after ground truth optimisation compared to the algorithmic output, against the ground truth. The optimisation step significantly improves algorithmic performance in finding strong registration correspondences.}
\renewcommand{\arraystretch}{1.4}
\setlength{\tabcolsep}{5pt}
\small
\begin{tabular}{c|cccc||cccc|cccc|cccc}
 & \multicolumn{4}{c||}{Synthetic}   & \multicolumn{4}{c|}{KITTI @ \SI{1.6}{\metre}} & \multicolumn{4}{c|}{KITTI @ \SI{11.3}{\metre}}  & \multicolumn{4}{c}{KITTI @ \SI{24}{\metre}}   \\ 
\multicolumn{1}{c|}{Method} & \multicolumn{1}{c}{F1} & \multicolumn{1}{c}{A} & \multicolumn{1}{c}{P} & \multicolumn{1}{c||}{R} & \multicolumn{1}{c}{F1} & \multicolumn{1}{c}{A} & \multicolumn{1}{c}{P} & \multicolumn{1}{c|}{R} & \multicolumn{1}{c}{F1} & \multicolumn{1}{c}{A} & \multicolumn{1}{c}{P} & \multicolumn{1}{c|}{R} & \multicolumn{1}{c}{F1} & \multicolumn{1}{c}{A} & \multicolumn{1}{c}{P} & \multicolumn{1}{c}{R} \\ \hline
\texttt{P2P-ICP} & 0.28 & 0.63 & 0.26 & 0.30 & 0.51 & 0.75 & 0.50 & 0.53 & 0.14 & 0.55 & 0.13 & 0.16 & 0.15 & 0.56 & 0.14 & 0.18 \\
\texttt{NAR-P2Pv1$^+$}  & \textbf{0.99} & \textbf{0.99} & \textbf{0.98} & \textbf{0.99} & \textbf{0.81} & \textbf{0.91} & \textbf{0.81} & \textbf{0.82} & \textbf{0.96} & \textbf{0.98} & \textbf{0.95} & \textbf{0.97} & \textbf{0.96} & \textbf{0.99} & \textbf{0.95} & \textbf{0.97} \\\hline
\texttt{P2P-ICP} & 0.28 & 0.63 & 0.26 & 0.30 & 0.51 & 0.75 & 0.50 & 0.53 & 0.14 & 0.55 & 0.13 & 0.16 & 0.15 & 0.56 & 0.14 & 0.18 \\
\texttt{NAR-P2Pv2$^+$} & \textbf{1.00} & \textbf{1.00} & \textbf{1.00} & \textbf{1.00} & \textbf{0.76} & \textbf{0.89} & \textbf{0.74} & \textbf{0.79} & \textbf{0.94} & \textbf{0.97} & \textbf{0.92} & \textbf{0.95} & \textbf{0.96} & \textbf{0.98} & \textbf{0.95} & \textbf{0.97} \\ \hline
\texttt{P2L-ICP} & 0.04 & 0.50 & 0.03 & 0.06 & 0.49 & 0.73 & 0.48 & 0.50 & 0.06 & 0.51 & 0.05 & 0.08 & 0.05 & 0.50 & 0.04 & 0.07 \\
\texttt{NAR-P2L$^+$} & \textbf{0.99} & \textbf{1.00} & \textbf{0.99} & \textbf{1.00} & \textbf{0.84} & \textbf{0.92} & \textbf{0.84} & \textbf{0.85} & \textbf{0.89} & \textbf{0.95} & \textbf{0.87} & \textbf{0.91} & \textbf{0.94} & \textbf{0.98} & \textbf{0.92} & \textbf{0.96} \\ \hline
\texttt{G-ICP} & 0.16 & 0.56 & 0.15 & 0.18 & 0.39 & 0.68 & 0.38 & 0.40 & 0.08 & 0.52 & 0.07 & 0.10 & 0.11 & 0.53 & 0.10 & 0.12 \\
\texttt{NAR-GICP$^+$} & \textbf{0.99} & \textbf{1.00} & \textbf{0.99} & \textbf{0.99} & \textbf{0.80} & \textbf{0.90} & \textbf{0.78} & \textbf{0.82} & \textbf{0.89} & \textbf{0.96} & \textbf{0.86} & \textbf{0.92} & \textbf{0.87} & \textbf{0.94} & \textbf{0.85} & \textbf{0.90}
\end{tabular}
\vspace{-5pt}

\label{tab:classification}
\end{table*}

\textbf{Ground Truth Optimisation. }
A key advantage of NAR-*ICP over the algorithms is that we can further improve their performance by adding a ground truth optimisation step. We note that this step is not part of the original NAR implementation in CLRS-30. We assess its effectiveness by comparing its performance not only against \texttt{NAR-CLRS}, but also against the baseline algorithms, evaluating the predicted outputs after optimisation against the algorithms' outputs. Our results in \Cref{tab:mse_optimisation}, \Cref{tab:classification}, and earlier in \Cref{tab:y_gt_rte_rre}, demonstrate that the optimised NAR-*ICP models consistently outperform the benchmarks across all datasets, both in predicting the ground truth transformed point clouds and in identifying the correct correspondences. This is evident by the significantly lower \gls{mse}$^{\textbf{GT}}$ scores and substantially improved classification metrics, F1$^{\textbf{GT}}$, P$^{\textbf{GT}}$, R$^{\textbf{GT}}$, and A$^{\textbf{GT}}$. Additionally, our optimised version achieves notably improved predictive performance against the training signal compared to \texttt{NAR-CLRS} in: a) \Cref{tab:mse_pred} and \Cref{tab:class_pred}, where the final output of NAR-*ICP is calculated equivalently to \texttt{NAR-CLRS} but with our learned termination, and b) in \cref{tab:terminations}, where NAR-*ICP is executed without our learned termination, in \texttt{NAR-CLRS}.

\section{Ablation}
\label{sec:ablation}
To find the optimal processor architecture, hints configuration, and teacher-forcing probability, we compared the performance of various configurations on the metrics defined in \cref{sec:metrics}, using the synthetic dataset and \texttt{P2P-ICP} as the baseline.

\begin{figure}[h]
    \centering
        \includegraphics[width=0.8\columnwidth]{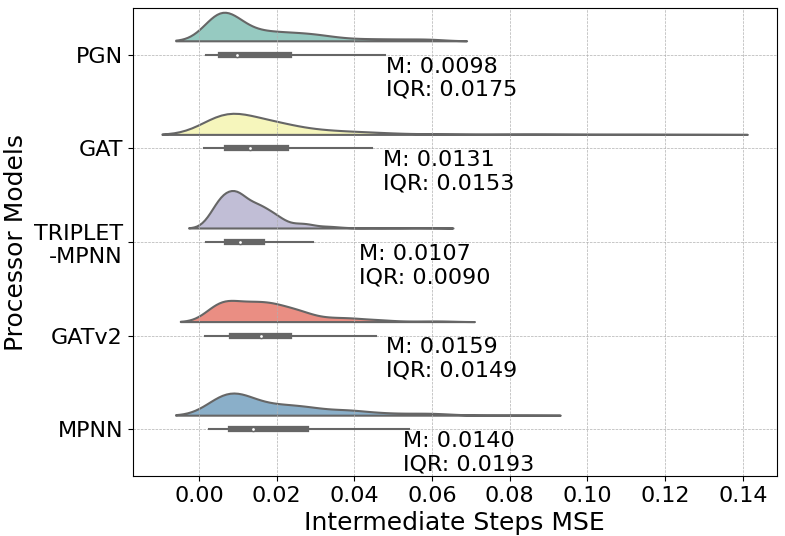}
        \caption{Ablation study of different model architectures, comparing the MSE$^{\textbf{t}}(\downarrow)$ error distributions of intermediate steps predictions.}
    \label{fig:ablation_model}
\end{figure}

\textbf{Processors. }
We tested the following processors on the synthetic dataset using the CLRS-30 Benchmark \cite{clrs}:
(a) Pointer-Graph Network (PGN) \cite{pgn}, (b) Graph Attention Network (GAT) \cite{gat}, (c) Triplet-Message-Passing Neural Network (MPNN) \cite{generalistNAR}, (d) GATv2 \cite{brody2022how}, and (e) MPNN \cite{mpnn}. 
Our results in \Cref{tab:ablation_model} demonstrate that the MPNN-based models performed best, with the Triplet-MPNN mostly outperforming the MPNN. As expected, the Triplet-MPNN performed better at also predicting the correspondences in both cases. By comparing the median and IQR in \Cref{fig:ablation_model}, we concluded that the Triplet-MPNN achieved the best performance trade-off in predicting the intermediate steps, as indicated by its low median and IQR along with the very small number of outliers.

\setlength{\tabcolsep}{4pt}
\begin{table}[h]
\centering
\small
\renewcommand{\arraystretch}{1.4}
\setlength{\tabcolsep}{5pt}
        \caption{Ablation study of different model architectures, comparing average MSE $(\downarrow)$ scores for the predicted transformed point clouds, along with F1 $(\uparrow)$ and balanced accuracy A $(\uparrow)$ for the predicted correspondences.}
\begin{tabular}{l|ccc||ccc}
                   & \multicolumn{3}{c||}{\texttt{NAR-P2Pv2$^+$}}       & \multicolumn{3}{c}{\texttt{NAR-P2Pv2}} \\ 
\multirow{-2}{*}{} & \multicolumn{1}{c}{MSE$^{\textbf{GT}}$} & \multicolumn{1}{c}{F1$^{\textbf{GT}}$} & \multicolumn{1}{c||}{A$^{\textbf{GT}}$} & \multicolumn{1}{c}{MSE$^{\textbf{T}}$} & \multicolumn{1}{c}{F1$^{\textbf{T}}$} & \multicolumn{1}{c}{A$^{\textbf{T}}$} \\ 
\hline
PGN   & 0.010    & 0.62    & 0.84       & 0.050     & 0.04   & 0.49    \\

GAT-Full  & 0.008   & 0.56   & 0.81   & 0.045   & 0.22 & 0.59   \\
Triplet-MPNN       & 0.007                   & \textbf{0.99}                  & \textbf{0.99}     & \textbf{0.017}                   & \textbf{0.30}                  & \textbf{0.63} \\

GATv2-Full    & 0.009   & 0.54   & 0.80     & 0.035    & 0.24    & 0.60  \\

MPNN    & \textbf{0.002}      & 0.82   & 0.93    & 0.038       & 0.26    & 0.60    

\end{tabular}
\label{tab:ablation_model}
\end{table}

\begin{figure}[h]
    \centering
        \includegraphics[width=0.8\columnwidth]{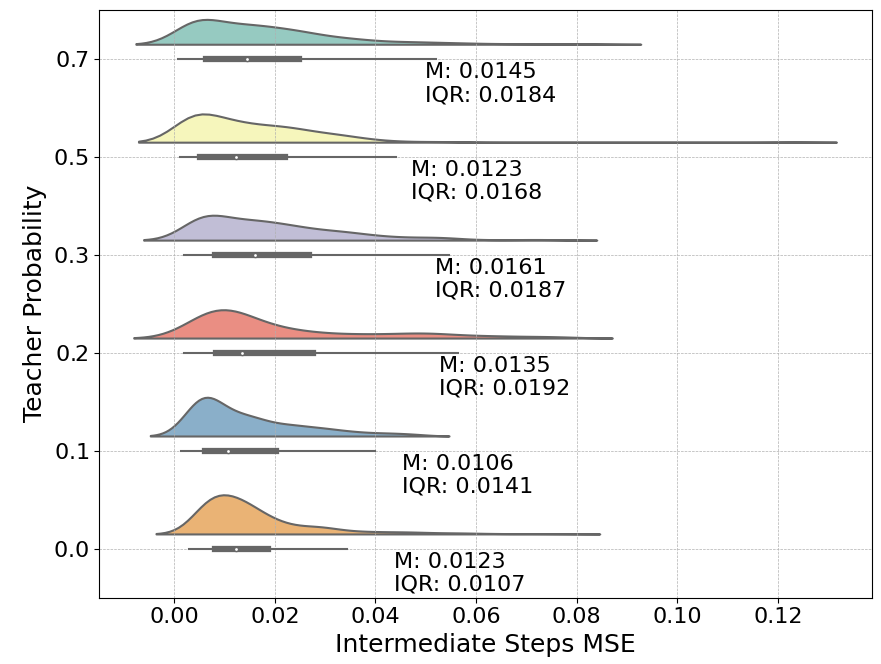}
        \caption{Ablation study of different teacher probability values, comparing the MSE$^{\textbf{t}}(\downarrow)$ error distributions of intermediate steps predictions.}
    \label{fig:ablation_teacher}   

\end{figure}

\textbf{Teacher-Forcing Probability. } Given the nature of the teacher-forcing optimisation, we are primarily interested in the model's performance at predicting the transformed point clouds at each intermediate step. To select the optimal probability threshold, $P_T$, we compared different threshold values against the model's output. Comparing the model's predictive performance for each, as seen in \Cref{fig:ablation_teacher}, we concluded that the optimal probability threshold is $P_T=0.1$ as it achieves the lowest median, the second-lowest IQR, and the fewest outliers. 

\textbf{Hint Configuration. }We tested various hint configurations to determine the optimal setup for capturing the trajectory of the algorithm. Each configuration varies in the selection of intermediate algorithmic steps that are passed as hints to the processor.
We evaluated our model's performance against the following hint configurations:
\begin{enumerate}[itemsep=0pt]
    \item \texttt{P2}: Pass only the output of the second phase of the algorithm.
    \item \texttt{P12}: Pass the output of both phases of the algorithm.
    \item \texttt{P1I}: Pass the intermediate calculations of the first phase of the algorithm - i.e. the correspondence-finding iterations.
    \item \texttt{P1I2}: Pass the intermediate calculations of the first phase of the algorithm along with the output of the second phase.
\end{enumerate}

\setlength{\tabcolsep}{4pt}
\begin{table}[h]
\centering
\small
\renewcommand{\arraystretch}{1.4}
\setlength{\tabcolsep}{7pt}
        \caption{Ablation study of different hint configurations for baseline \texttt{NAR-P2P}. \textbf{Bold} and \underline{underline} values indicate the best and second-best results, respectively. We compare average MSE$(\downarrow)$ for the predicted transformed point clouds, along with F1$(\uparrow)$ and balanced accuracy A$(\uparrow)$ for the predicted correspondences.}
\begin{tabular}{l|ccc||ccc}
                   & \multicolumn{3}{c||}{\texttt{NAR-P2P$^+$}}    & \multicolumn{3}{c}{\texttt{NAR-P2P}} \\ 
\multirow{-2}{*}{} & \multicolumn{1}{c}{MSE$^{\textbf{GT}}$} & \multicolumn{1}{c}{F1$^{\textbf{GT}}$} & \multicolumn{1}{c||}{A$^{\textbf{GT}}$} & \multicolumn{1}{c}{MSE$^{\textbf{T}}$} & \multicolumn{1}{c}{F1$^{\textbf{T}}$} & \multicolumn{1}{c}{A$^{\textbf{T}}$} \\ \hline
\texttt{P2}          & \underline{0.011}                   & 0.93                  & 0.97         & \textbf{0.029}                   & 0.26                  & \underline{0.61}                     \\
\texttt{P12}    & \textbf{0.008}                   & \underline{0.95}                  & \underline{0.98}      & \underline{0.049}                   & 0.21                  & 0.58                     \\
\texttt{P1I}          & 0.016                   & \textbf{1.00}                  & \textbf{1.00}     & 0.09                   & \underline{0.35}                  & \underline{0.61}         \\
\texttt{P1I2}      & 0.014                   & \textbf{1.00}                  & \textbf{1.00}     & 0.058                   & \textbf{0.43}                  & \textbf{0.66}                     \\

\end{tabular}

\label{tab:ablation_hints}

\end{table}

Here, we initially evaluated the final predictions from the models, as shown in \Cref{tab:ablation_hints}. The results demonstrate that different hint configurations achieve comparable performance in predicting the final output of the algorithm and the ground truth, with \texttt{P2} and \texttt{P12} outperforming the others in predicting the transformed point clouds, as indicated by lower average MSE scores. Our experiments indicate that to effectively learn the intermediate representations of the algorithms, hints should be representative of the algorithm's entire trajectory. However, when the changes between the hints are minimal and the inputs are imbalanced -- such as in the cases of \texttt{P1I} and \texttt{P1I2} -- the model's performance suffers in predicting the intermediate hints. For configurations \texttt{P1I} and \texttt{P1I2}, the intermediate ground truth hints are mostly repetitive, and the changes between scalar hints are minimal. Therefore, we concluded that the configurations \texttt{P2} and \texttt{P12} are of most interest and, as such, they have both been explored further above in \texttt{NAR-P2Pv1} and \texttt{NAR-P2Pv2}, respectively.

\section{Integration of NAR-*ICP in Learning Pipelines \label{sec:learning_pipeline}}

This section describes how NAR-*ICP can be used as a sub-component of a larger learning pipeline. To demonstrate its successful integration, we developed a contrastive-learning framework that generates latent feature representations of objects in SemanticKITTI \cite{semantickitti}. These learned features are then used in the input of NAR-*ICP, demonstrating its flexibility in handling diverse input types.

\begin{figure}[]
    \centering
    \captionsetup{justification=centering}
    \begin{subfigure}{0.8\columnwidth}
    \centering
    \includegraphics[trim=290 250 200 250,clip,width=\columnwidth]{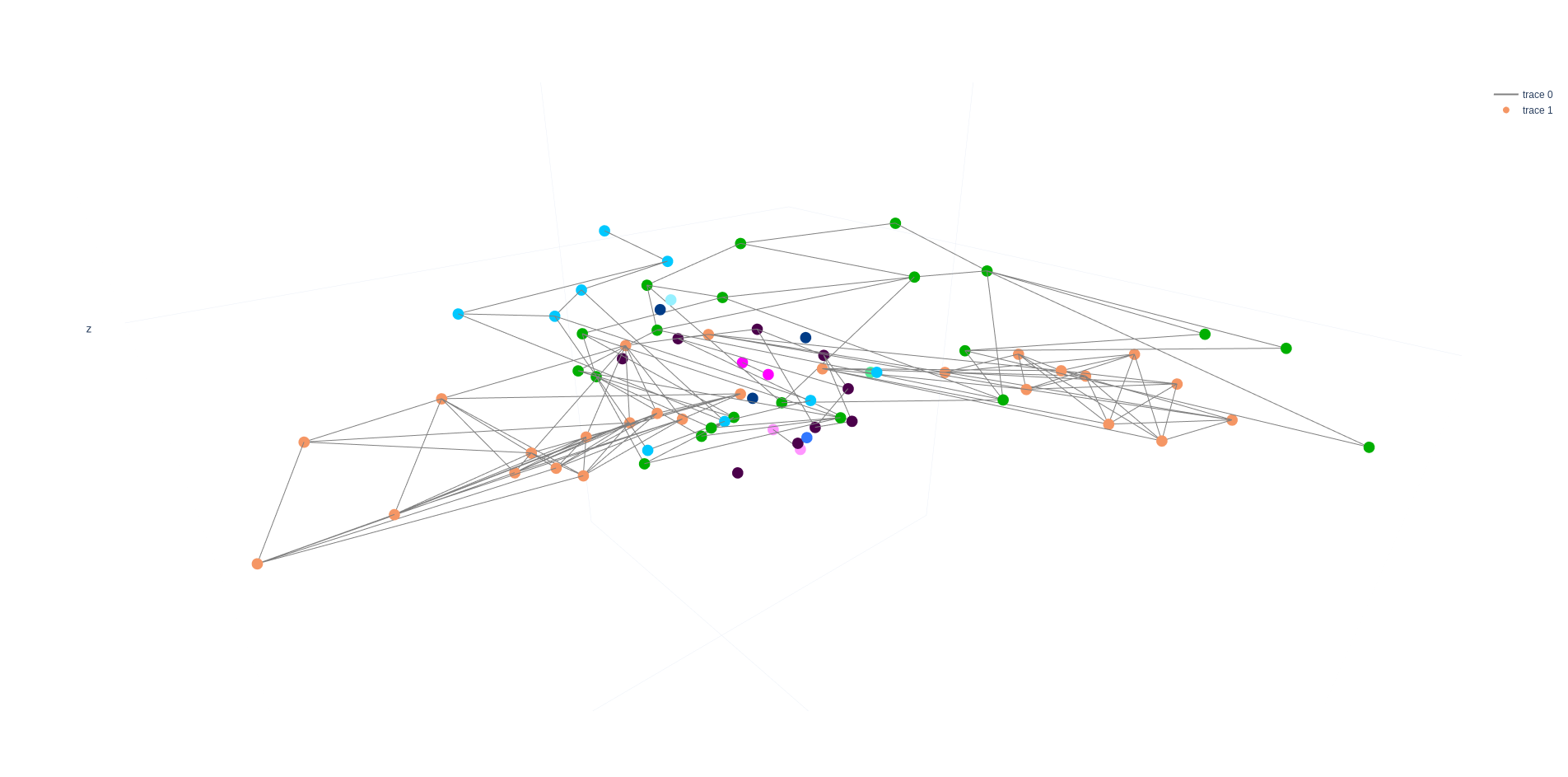}
    \caption{}
    \end{subfigure}
    \begin{subfigure}{0.8\columnwidth}
    \centering
    \includegraphics[trim=460 150 140 200,clip,width=\columnwidth]{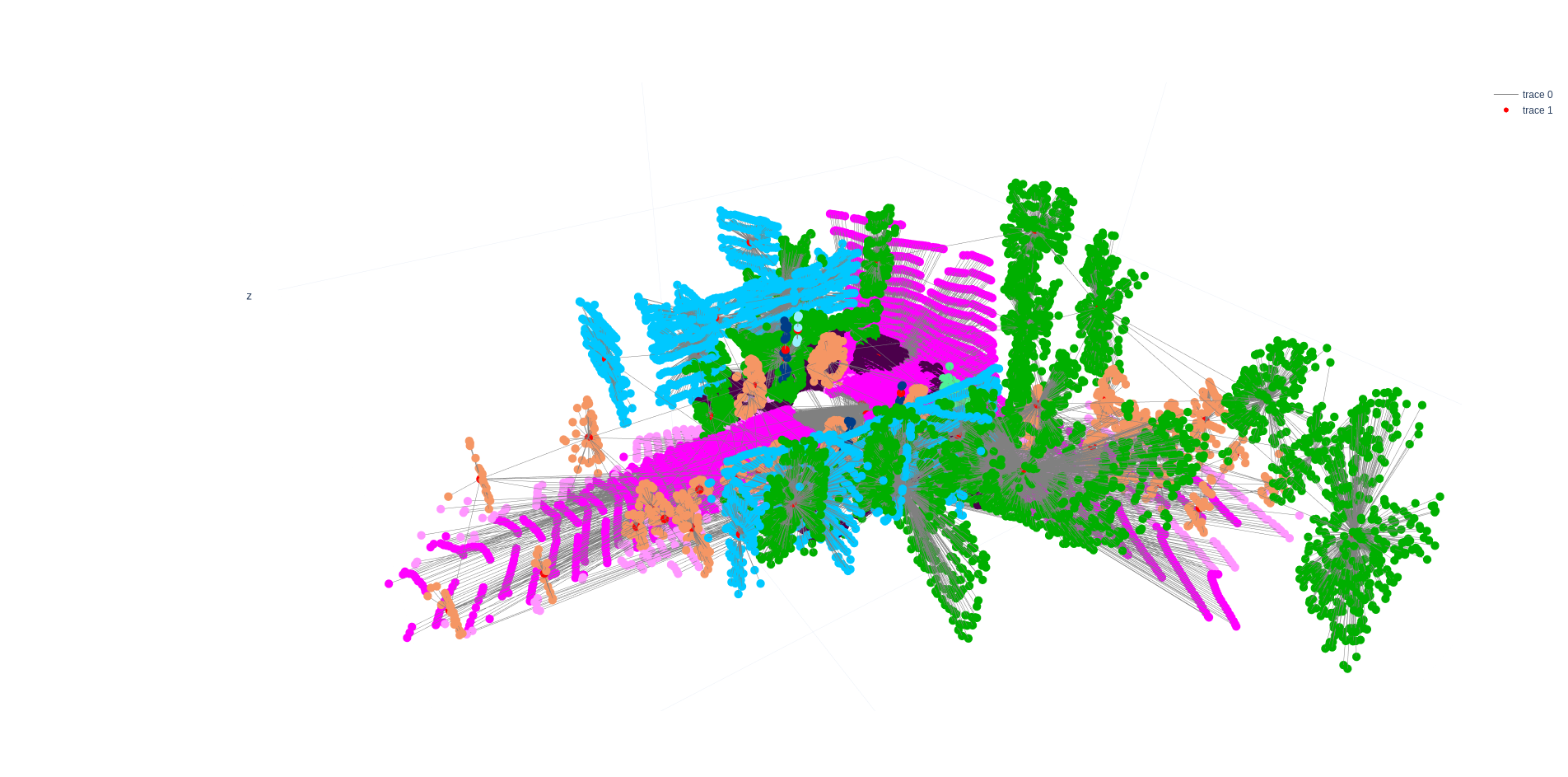}
    \caption{}
    \end{subfigure}
    \captionsetup{justification=justified}
    \caption{Example of a scan from the real-world dataset, with the centroids in (a) and the corresponding point clouds in (b), where each object's sub-point cloud is coloured differently.
    \label{fig:datasets}}
\end{figure}

To achieve this, we learn a useful embedding representation for each object in the real-world dataset, leveraging SimCLR \cite{simclr} along with a \gls{gnn} to generate strong embedding representations for each object, which are then used as input $x_i^{(t)}$ to our approach. In particular, during training, while the underlying ICP algorithm is solved using the original Cartesian data of the centroids to extract the trajectory of the algorithm, the NAR-*ICP encoder receives the latent features as input graph nodes. Contrastive learning is particularly effective in distinguishing between similar and dissimilar data points, facilitating robust feature learning. \Cref{fig:datasets} illustrates an example of a point cloud used in this contrastive-learning-based approach, depicting both the centroids used for registration and the points associated with each object. Our results in \cref{tab:contrastive_rte_rre} and \cref{tab:contrastive_mse_f1} demonstrate that our combined contrastive and \gls{nar} framework achieves performance comparable to NAR-*ICP trained on Cartesian data, highlighting the ease of integrating NAR-*ICP into training pipelines.

\begin{table}[h]
\centering
\renewcommand{\arraystretch}{1.4}
\small
\setlength{\tabcolsep}{8pt}
\caption{Registration performance of the integrated learning pipeline for the final step of NAR-*ICP models compared to the final step of the algorithm and the ground truth in RTE$^{\textbf{T}}(\downarrow)$, RRE$^{\textbf{T}}(\downarrow)$ and RTE$^{\textbf{GT}}(\downarrow)$, RRE$^{\textbf{GT}}(\downarrow)$, respectively.}
\begin{tabular}{l|cc||cc}
 Method & RTE$^{\textbf{T}}$ & RRE$^{\textbf{T}}$ & RTE$^{\textbf{GT}}$ & RRE$^{\textbf{GT}}$ \\ \hline
\texttt{NAR-P2Pv1} & 0.818 & 1.384 & 1.014 & 1.998 \\
\texttt{NAR-P2Pv2} & 0.821 & 1.442 & 1.167 & 2.127 \\
\texttt{NAR-P2L} & 1.208 & 2.201 & 0.783 & 1.338  \\
\texttt{NAR-GICP} & 0.881 & 1.644 & 0.935 & 2.016 
\end{tabular}

\label{tab:contrastive_rte_rre}
\end{table}

\begin{table}[h]
\centering
\renewcommand{\arraystretch}{1.4}
\small
\setlength{\tabcolsep}{6pt}
\caption{Prediction performance of the integrated learning pipeline for the NAR-*ICP models after ground truth optimisation, evaluated using RTE$^{\textbf{GT}}(\downarrow)$ and RRE$^{\textbf{GT}}(\downarrow)$ for registration performance, average \gls{mse}$^{\textbf{GT}}$$(\downarrow)$ for the predicted transformed point clouds, and F1$^{\textbf{GT}}(\uparrow)$ and balanced accuracy A$^{\textbf{GT}}(\uparrow)$ for correspondences.}
\begin{tabular}{c|cc||ccc}
Method & RTE & RRE & $MSE$ & F1 & A \\ \hline
\texttt{NAR-P2Pv1$^+$} & 0.555 & 1.004 & \textbf{0.012} & \textbf{0.64} & \textbf{0.82} \\
\texttt{NAR-P2Pv2$^+$} & 0.596 & 1.058 & 0.023 & \textbf{0.64} & \textbf{0.82} \\
\texttt{NAR-P2L$^+$} & 0.621 & 1.151 & 0.026 & 0.54 & 0.78 \\
\texttt{NAR-GICP$^+$} & \textbf{0.507} & \textbf{0.937} & 0.019 & 0.63 & 0.82
\end{tabular}
\label{tab:contrastive_mse_f1}
\end{table}

Registration algorithms work well with 3D point clouds, while high-level semantics and learned embeddings are incorporated into learned methods but not directly into the calculations. This integration experiment is interesting as it presents complex, abstract data in the input of the neural execution and demonstrates that we can incorporate semantics and latent features to solve ICP-based algorithms. This framework is used to demonstrate the flexibility and generalisability of our method as well as its usefulness as a fully differential component of a larger learning system.


\section{Conclusion and Future Directions}
\label{sec:conclusion}
This work proposes a novel \gls{nar}-based approach for learning to approximate the intermediate steps of ICP-based algorithms, introducing the framework into the field of robotics. Our method not only mimics point cloud registration pipelines but also consistently outperforms them. We additionally demonstrate the reliability and flexibility of NAR-*ICP when handling noisy and complex inputs. The NAR framework is further extended by leveraging the specific architecture of the ICP-based algorithms, enhancing its functionality and showcasing its effectiveness in approximating complex multi-step algorithms. Our method aims to advance current robotics systems by proposing a more interpretable and efficient learning paradigm. By integrating classical algorithms with deep learning, we combine structured reasoning and logical computations with the adaptability and generalisation capabilities of neural networks. 
\gls{nar}-based models provide access to their intermediate computations, leading to more reliable, robust, and transparent robotics systems. 
As such, we foresee that learning in the space of autonomous navigation, planning, and manipulation will reveal interesting applications of \gls{nar}, extending NAR-*ICP and the CLRS-30 Benchmark further. Furthermore, the inherent interpretability of our method can enhance human-robot interaction by enabling systems to explain their decision-making process, leading to more transparent and reliable systems for real-world applications.




\bibliographystyle{IEEEtran}
\bibliography{NAR}

\end{document}